\documentclass[preprint,12pt]{elsarticle}

\usepackage{amsmath,amsfonts}
\usepackage{algorithmic}
\usepackage{array}
\usepackage[tight,footnotesize]{subfigure}
\usepackage{textcomp}
\usepackage{stfloats}
\usepackage{url}
\usepackage{verbatim}
\usepackage{graphicx}
\def\BibTeX{{\rm B\kern-.05em{\sc i\kern-.025em b}\kern-.08em
    T\kern-.1667em\lower.7ex\hbox{E}\kern-.125emX}}
\usepackage{balance}
\usepackage{booktabs}
\usepackage{caption}
\usepackage{makecell}
\usepackage{adjustbox}
\usepackage{wrapfig}
\usepackage[linesnumbered,ruled]{algorithm2e}
\usepackage{amssymb}
\usepackage{lineno}
\usepackage{float}
\usepackage{algorithmic}
\usepackage[tight,footnotesize]{subfigure}
\usepackage{booktabs}
\usepackage{caption}
\usepackage{makecell}
\usepackage{adjustbox}
\usepackage{wrapfig}
\usepackage{xcolor}
\usepackage[linesnumbered,ruled]{algorithm2e}
\usepackage{listings}
\usepackage{makecell}

\usepackage{tikz}
\usetikzlibrary{shapes.geometric, arrows.meta, positioning}

\usepackage{natbib}

\usepackage{pifont}
\newcommand{\cmark}{\ding{51}}%
\newcommand{\xmark}{\ding{55}}%

\journal{Journal Name}

\begin{document}
\sloppy
\setlength{\parskip}{0pt}

\begin{frontmatter}


\title{Towards an Agent-First Web: Redesigning the Web for AI Agents}

\author[label1]{Eranga Bandara}
\ead{cmedawer@odu.edu}

\author[label1]{Ross Gore}
\ead{rgore@odu.edu}

\author[label1]{Ravi Mukkamala}
\ead{rmukkama@odu.edu}

\author[label10]{Asanga Gunaratna}
\ead{asanga.gunaratna@complianceoslab.app}

\author[label1]{Safdar H. Bouk}
\ead{sbouk@odu.edu}

\author[label4]{Xueping Liang}
\ead{xuliang@fiu.edu}

\author[label1]{Peter Foytik}
\ead{pfoytik@odu.edu}

\author[label3]{Abdul Rahman}
\ead{abdulrahman@deloitte.com}

\author[label7]{Sachini Rajapakse}
\ead{sachini.rajapakse@iciclelabs.ai}

\author[label15]{Isurunima Kularathna}
\ead{nima@linesandloops.art}

\author[label7]{Pramoda Karunarathna}
\ead{pramoda.karunarathna@iciclelabs.ai}

\author[label7]{Chalani Rajapakse}
\ead{chalani.rajapakse@iciclelabs.ai}

\author[label5]{Ng Wee Keong}
\ead{awkng@ntu.edu.sg}

\author[label6]{Kasun De Zoysa}
\ead{kasun@ucsc.cmb.ac.lk}

\author[label19]{Tharaka Hewa}
\ead{tharaka.hewa@oulu.fi}

\author[label8]{Amin Hass}
\ead{amin.hassanzadeh@accenture.com}

\author[label17]{Wathsala Herath}
\ead{wathsala.herath@agentsway.ai}

\author[label20]{Aruna Withanage}
\ead{aruna@effectz.ai}

\author[label20]{Nilaan Loganathan}
\ead{nilaan@effectz.ai}

\author[label14]{Atmaram Yarlagadda}
\ead{atmaram.yarlagadda.civ@health.mil}

\author[label1]{Sachin Shetty}
\ead{sshetty@odu.edu}

\address[label1]{Old Dominion University, Norfolk, VA, USA}
\address[label10]{AI Motion Labs, Melbourne, Australia}
\address[label3]{Deloitte \& Touche LLP, USA}
\address[label4]{Florida International University, USA}
\address[label7]{IcicleLabs.AI}
\address[label15]{Linesandloops.art}
\address[label8]{Accenture Technology Labs, Arlington, VA, USA}
\address[label5]{Nanyang Technological University, Singapore}
\address[label6]{University of Colombo, Sri Lanka}
\address[label17]{Agentsway.AI}
\address[label20]{Effectz.AI}
\address[label19]{Center for Wireless Communications, University of Oulu, Finland}
\address[label14]{McDonald Army Health Center, Newport News, VA, USA}

\begin{abstract}
The World Wide Web was architected on a foundational assumption that 
has held for three decades: that the primary consumer of web content 
is a human being. This assumption permeates every layer of the web — 
its access model presumes human visitors, its economic foundations 
rest on human attention, and its content architecture targets human 
perception. The rapid emergence of AI agents as primary intermediaries 
between humans and web content fundamentally invalidates this 
assumption. Today, AI agents browse, synthesize, and act on web 
content on behalf of humans at scale — yet the web actively resists 
them through blanket blocking, CAPTCHA-based exclusion, and economic 
models that treat agent access as extraction rather than legitimate 
interaction.

This paper proposes a principled redesign of the web across three 
interdependent layers. At the \textit{access layer}, we argue that 
agents acting on behalf of humans should inherit equivalent access 
rights to those humans — governed not by blanket blocking but by 
rate limiting and standardized agent identification metadata carried 
in HTTP requests, analogous to how browsers currently identify 
themselves to servers. We propose a dual-layer web architecture in 
which human-readable content and agent-optimized content coexist 
during a transition period, with a phased migration path toward an 
agent-first publishing standard. At the \textit{economic layer}, we 
reject universal pay-per-query models in favor of an intent-based 
tier framework grounded in the principle that an agent's economic 
obligation mirrors that of the human it represents — the 
\textit{agent-as-human-proxy principle}. We propose a token-based 
subscription model in which content access is metered in tokens 
rather than pageviews, alongside a commissioned content economy in 
which human intentionality anchors AI-generated content production 
and breaks the self-reinforcing generative loop. At the 
\textit{content layer}, we identify a structural threat we term 
\textit{epistemic recursion} — the self-referential loop in which 
AI-generated content is consumed by AI agents to produce further 
content, progressively detaching web knowledge from human ground 
truth. To counter this, we propose the Agent Text Markup Language 
(ATML) — a semantic content format optimized for agent consumption 
— alongside a four-level human supervision tier model and a 
cryptographic provenance chain architecture that makes the degree 
of human oversight machine-readable and verifiable.

Together, these three layers constitute a framework of ten design 
principles for an agent-first internet — one in which agents are 
not threats to be blocked but first-class citizens whose integration 
requires renegotiating the web's foundational social contract across 
access, economics, and content.
\end{abstract}


\end{frontmatter}

\section{Introduction}

The World Wide Web, since its inception by Berners-Lee in 1989, was 
designed with a singular assumption: that the primary consumer of 
web content is a human being \cite{berners1989information}. This 
assumption permeates every layer of the web's architecture. Hypertext 
Markup Language (HTML) renders content visually for human perception 
\cite{raggett1999html}. Search engines optimize content discovery 
around human attention and intent \cite{brin1998anatomy}. Economic 
models monetize human engagement through advertising impressions and 
clicks \cite{evans2008economics}. Security mechanisms such as CAPTCHAs 
are explicitly designed to distinguish humans from automated agents 
\cite{von2003captcha}. For three decades, this human-centric 
architecture served the web well — enabling an unprecedented global 
information ecosystem that transformed commerce, communication, and 
knowledge.

This foundational assumption is now breaking down. The rapid 
advancement of large language models (LLMs)~\cite{llm, aitrust-os} and autonomous AI agents 
has introduced a fundamentally new class of web participant~\cite{agentic-ai-transition-organization, agentic-workflow-practicle-guide} — one 
that browses, synthesizes, and acts on web content not for itself, 
but as a proxy for a human user \cite{wang2024survey}. Unlike 
traditional web crawlers, which indexed content to route human 
traffic back to publishers, modern AI agents complete tasks 
end-to-end — booking flights, answering research questions, drafting 
documents — without the human ever visiting the underlying web 
resource \cite{yao2023webagents}. The web's human-centric 
architecture was not designed for this interaction model, and the 
friction is now measurable and severe.

Empirically, this tension manifests across all three layers of the 
web. At the access layer, infrastructure providers have begun blocking 
AI agents by default. Cloudflare, which routes over 16\% of global 
internet traffic, moved in July 2025 to block AI crawlers unless 
explicitly permitted, introducing a pay-per-crawl model that treats 
agent access as a commercial transaction rather than a default right 
\cite{cloudflare2025block}. The crawl-to-referral ratio of AI agents 
reveals the economic asymmetry driving this response: by mid-2025, 
Anthropic's crawler exhibited a ratio of 73,000 crawled pages for 
every single human referral generated — fundamentally breaking the 
value exchange that historically justified open crawler access 
\cite{cloudflare2025ratio}. At the economic layer, zero-click 
searches now account for approximately 60\% of all Google queries, 
rising to 93\% in AI-native search modes \cite{semrush2025zeroclicks}, 
with click-through rates at position one falling from 27\% to 11\% 
\cite{sistrix2026ctr}. Publishers report organic traffic declines of 
70--80\%, with some characterizing the shift as an extinction-level 
event for web-dependent media \cite{sistrix2026ctr}. At the content 
layer, a deeper structural problem emerges: as AI agents increasingly 
consume AI-generated content to produce further content, the web 
risks entering a self-referential generative loop that progressively 
erodes its epistemic foundations \cite{shumailov2023curse} — a 
phenomenon we term \textit{epistemic recursion}.

The web's response to these pressures has been reactive and 
fragmented. Infrastructure providers block agents by default 
\cite{cloudflare2025block}. Protocol working groups propose 
interoperability standards in isolation \cite{anthropic2024mcp, 
google2025a2a, microsoft2025nlweb}. Economic responses treat agent 
access as a billing problem rather than a design problem 
\cite{cloudflare2025block, perplexity2024sharing}. Content provenance 
efforts address detection of AI-generated content without addressing 
the architectural conditions that produce epistemic recursion 
\cite{c2pa2024, kirchenbauer2023watermark}. Critically, no existing 
work addresses all three layers together, nor frames the challenge 
as what we argue it fundamentally is: a failure of the web's 
foundational social contract that requires principled redesign, not 
reactive patches~\cite{mcc, agentsway}.

This paper makes the following argument: the web requires simultaneous 
redesign across three interdependent layers — access, economics, and 
content — grounded in a single philosophical anchor: that AI agents 
acting on behalf of humans are first-class web citizens, entitled to 
the same presumption of access as the humans they represent, subject 
to equivalent obligations, and deserving of an architectural 
environment designed for their interaction model rather than one that 
merely tolerates them.

Concretely, we propose three mechanism clusters. At the access 
layer, we propose \textit{agent identification metadata} — 
standardized HTTP request headers that allow agents to declare their 
identity, the human they represent, and their intent, analogous to 
the User-Agent and Accept headers browsers currently use 
\cite{fielding1999http} — alongside \textit{agents.txt}, a 
machine-readable access policy standard that replaces the inadequate 
\texttt{robots.txt} honor system with graduated, intent-aware access 
declarations. Together these enable servers to apply rate limiting 
rather than blanket blocking, and to serve agent-optimized content 
via a \textit{dual-layer web architecture} that supports gradual 
migration rather than disruptive replacement. At the economic layer, 
we propose a \textit{token-based subscription model} in which content 
access is metered in tokens rather than pageviews — directly 
compatible with existing AI API pricing infrastructure — alongside 
a \textit{commissioned content economy} in which human 
intentionality anchors AI content production, and a free tier 
governed by rate limits mirroring the open source / proprietary 
software distinction \cite{raymond1999cathedral}. At the content 
layer, we propose the \textit{Agent Text Markup Language (ATML)} 
— a semantic content format optimized for agent consumption — 
alongside a \textit{four-level human supervision tier model} and 
a \textit{cryptographic provenance chain architecture} that make 
the degree of human oversight machine-readable and verifiable, 
breaking the epistemic recursion loop at its structural root.

We make the following specific contributions:

\begin{itemize}
    \item We diagnose the web's human-centric design assumptions 
    across three layers and demonstrate empirically that each is 
    incompatible with agent-first interaction at scale 
    (Section~\ref{sec:diagnosis}).

    \item We propose \textit{agent identification metadata} and 
    \textit{agents.txt} as lightweight, backward-compatible 
    mechanisms for agent identification, intent declaration, and 
    access policy over existing HTTP infrastructure, enabling 
    rate limiting over blocking and dual-layer content serving 
    (Section~\ref{sec:access}).

    \item We introduce an \textit{intent-based economic tier model} 
    grounded in the agent-as-human-proxy principle, with a 
    token-based subscription mechanism and commissioned content 
    economy directly compatible with existing AI API infrastructure 
    (Section~\ref{sec:economics}).

    \item We introduce the concept of \textit{epistemic recursion} 
    — the self-referential loop in which AI-generated content is 
    consumed by AI agents to produce further content, progressively 
    detaching web knowledge from human ground truth — and propose 
    ATML, human supervision tiers, and a cryptographic provenance 
    chain as architectural responses (Section~\ref{sec:content}).

    \item We synthesize these contributions into a unified framework 
    of ten design principles for an agent-first internet, 
    positioning the challenge as a sociotechnical problem requiring 
    renegotiation of the web's foundational social contract 
    (Section~\ref{sec:framework}).
\end{itemize}

The remainder of this paper is structured as follows. 
Section~\ref{sec:related} reviews related work across agent 
protocols, web economics, and content architecture. 
Section~\ref{sec:diagnosis} diagnoses the three-layer failure of 
the human-centric web. Section~\ref{sec:access} presents the access 
layer redesign. Section~\ref{sec:economics} presents the economic 
layer redesign. Section~\ref{sec:content} presents the content layer 
redesign including ATML, supervision tiers, and epistemic recursion 
prevention. Section~\ref{sec:framework} synthesizes the unified 
framework and ten principles. Section~\ref{sec:challenges} discusses 
open challenges. Section~\ref{sec:conclusion} concludes.

\section{Related Work}

\label{sec:related}

Research relevant to redesigning the web for AI agents spans four 
broad areas: (1) AI agent architectures and web interaction, (2) 
agent communication protocols and infrastructure, (3) web economics 
and the attention economy, and (4) content provenance and epistemic 
integrity. While each area has produced significant contributions, 
no existing work addresses all three layers — access, economics, and 
content — simultaneously, nor frames the challenge as a renegotiation 
of the web's foundational social contract. Table~\ref{tab:relatedwork} 
summarizes how representative works relate to the key dimensions of 
our framework.

\subsection{AI Agent Architectures and Web Interaction}

Early work on web-based AI agents focused on task completion in 
constrained environments. \citet{yao2023webagents} introduced WebShop, 
demonstrating that language model agents could navigate product search 
and purchase flows, establishing a benchmark for grounded web 
interaction. \citet{zhou2023webarena} extended this with WebArena, a 
realistic multi-domain environment covering e-commerce, content 
management, and developer tools, revealing the significant gap between 
agent capability and real-world web complexity. \citet{deng2023mind2web} 
introduced Mind2Web, the first dataset for generalist web agents capable 
of following natural language instructions across diverse websites, 
highlighting the challenge of generalizing across the heterogeneous 
structure of the human-designed web.

More recent work has examined agentic behavior at scale. 
\citet{wang2024survey} surveyed LLM-based autonomous agents across 
planning, memory, and tool use dimensions, noting that web interaction 
represents one of the most complex and underspecified deployment 
environments. \citet{durante2024agent} examined multi-agent systems 
operating in open-ended environments, identifying trust, coordination, 
and resource access as primary bottlenecks. Critically, none of these 
works examine the web architecture itself as a site of intervention — 
they treat the existing web as a fixed environment and attempt to 
improve agent capability within it. Our work takes the complementary 
position: that the web architecture must change to accommodate agents, 
not only that agents must improve to navigate the existing web.

\subsection{Agent Communication Protocols and Infrastructure}

A parallel research and engineering effort has focused on protocol-level 
infrastructure for agent interaction. \citet{anthropic2024mcp} 
introduced the Model Context Protocol (MCP), an open standard enabling 
AI agents to connect to external tools and data sources through a 
unified interface, addressing the fragmentation of agent-to-tool 
communication. Google's Agent-to-Agent (A2A) protocol 
\cite{google2025a2a} extended this to agent-to-agent communication, 
enabling coordination across heterogeneous agent systems. IBM's Agent 
Communication Protocol (ACP) \cite{ibm2025acp} and the Agent Network 
Protocol (ANP) \cite{anp2025} addressed further coordination challenges 
at network scale. \citet{microsoft2025nlweb} proposed NLWeb, bringing 
natural language interfaces to websites and positioning every NLWeb 
endpoint as an MCP server — a significant step toward agent-readable 
web content.

\citet{kapoor2024infrastructure} provided the most comprehensive survey 
of agent infrastructure requirements to date, cataloging use cases, 
limitations, and open problems across memory, execution, and 
communication layers. \citet{li2025agenticweb} proposed a conceptual 
model for the agentic web across intelligence, interaction, and economic 
dimensions, identifying architectural challenges in protocol design and 
agent orchestration. The W3C Agent Web Community Group 
\cite{w3c2025agentwebcg} identified four core trends in agent-web 
interaction and three primary challenges including data silos, 
human-machine interface friction, and absence of standard protocols.

While these works make important protocol-level contributions, they 
address the access and interoperability problem in isolation. They do 
not examine the economic model that must accompany open agent access, 
nor the epistemic consequences of AI-generated content recursion. Our 
framework synthesizes the protocol layer with the economic and content 
layers to produce a unified redesign proposal.

\subsection{Web Economics and the Attention Economy}

The economic foundations of the web have been studied extensively in 
the context of human interaction. \citet{evans2008economics} 
characterized the web's advertising model as a two-sided market 
connecting publishers and advertisers through human attention as the 
mediating resource. \citet{wu2016attention} traced the historical 
development of the attention economy, demonstrating how successive 
communication technologies — radio, television, the web — converged on 
attention capture as the primary economic mechanism.

The disruption of this model by AI is increasingly documented 
empirically. \citet{cloudflare2025ratio} reported crawl-to-referral 
ratios of 73,000:1 for AI crawlers by mid-2025, demonstrating the 
complete breakdown of the value exchange that historically justified 
open crawler access. \citet{semrush2025zeroclicks} documented that 93\% 
of searches in AI-native modes end without a click, while 
\citet{sistrix2026ctr} reported click-through rate declines from 27\% 
to 11\% at position one in AI-augmented search. \citet{cloudflare2025block} 
described the infrastructure response to these dynamics — default 
blocking of AI crawlers and introduction of pay-per-crawl models — 
as the first large-scale attempt to impose economic structure on agent 
access.

Proposed economic responses have been largely reactive and incomplete. 
Perplexity's publisher revenue sharing \cite{perplexity2024sharing} and 
Cloudflare's pay-per-crawl model \cite{cloudflare2025block} address 
specific friction points without a principled framework. 
\citet{nguyen2025agenteconomy} examined micropayment models for agent 
content access but did not address the agent-as-human-proxy principle 
or the distinction between personal and commercial agent use. Our work 
contributes the intent-based economic tier model — grounded in the 
philosophical principle that an agent's economic obligation mirrors 
that of the human it represents — as a principled alternative to 
universal payment requirements.

\subsection{Content Provenance and Epistemic Integrity}

The question of content provenance has gained urgency with the 
proliferation of AI-generated content. \citet{zellers2019grover} 
demonstrated early that neural text generation could produce 
indistinguishable fake news, motivating provenance research. 
\citet{kirchenbauer2023watermark} proposed watermarking techniques 
for LLM outputs, enabling statistical detection of AI-generated text. 
The C2PA (Coalition for Content Provenance and Authenticity) standard 
\cite{c2pa2024} established cryptographic provenance for media content, 
providing a technical foundation for authorship declaration at the 
content layer.

Most directly relevant to our epistemic recursion argument, 
\citet{shumailov2023curse} demonstrated the phenomenon of model 
collapse — the progressive degradation of model quality when trained 
on AI-generated data — providing empirical grounding for our theoretical 
framing. \citet{martini2024provenance} examined provenance requirements 
for agentic content pipelines, and \citet{bender2021stochastic} 
characterized large language models as stochastic parrots — systems 
that recombine existing patterns without grounding in world experience 
— a characterization that motivates the human-intentionality anchor we 
propose.

Existing provenance work addresses detection and attribution of 
AI-generated content but does not address the systemic architectural 
question of how the web should be redesigned to preserve epistemic 
integrity at scale. Our provenance-anchored content architecture 
extends this work by proposing a web-level standard for declaring 
content origin, derivation chain, and human oversight level — going 
beyond detection to prevention of epistemic recursion.

\subsection{Summary and Positioning}

Table~\ref{tab:relatedwork} summarizes the coverage of representative 
related works across the six key dimensions of our framework: agent 
access model, behavioral contracts, economic framework, intent-based 
tiers, epistemic recursion, and provenance architecture. As the table 
shows, existing works address individual dimensions in isolation. 
Our framework is the first to address all six dimensions simultaneously, 
grounded in a unified social contract redesign philosophy.

\begin{table*}[!htb]
\centering
\caption{Comparison of related work across key dimensions of the 
agent-first web redesign framework. \cmark~=~addressed; 
\xmark~=~not addressed; $\sim$~=~partially addressed.}
\begin{adjustbox}{width=1\textwidth}
\label{tab:relatedwork}
\begin{tabular}{lcccccc}
\toprule
\thead{Work} & 
\thead{Agent\\access model} & 
\thead{Behavioral\\contracts} & 
\thead{Economic\\framework} & 
\thead{Intent-based\\tiers} & 
\thead{Epistemic\\recursion} & 
\thead{Provenance\\architecture} \\
\midrule

Yao et al.~\cite{yao2023webagents} & 
$\sim$ & \xmark & \xmark & \xmark & \xmark & \xmark \\

Zhou et al.~\cite{zhou2023webarena} & 
$\sim$ & \xmark & \xmark & \xmark & \xmark & \xmark \\

Deng et al.~\cite{deng2023mind2web} & 
$\sim$ & \xmark & \xmark & \xmark & \xmark & \xmark \\

Wang et al.~\cite{wang2024survey} & 
$\sim$ & \xmark & \xmark & \xmark & \xmark & \xmark \\

Kapoor et al.~\cite{kapoor2024infrastructure} & 
\cmark & \xmark & $\sim$ & \xmark & \xmark & \xmark \\

Li et al.~\cite{li2025agenticweb} & 
\cmark & \xmark & $\sim$ & \xmark & \xmark & \xmark \\

Anthropic MCP~\cite{anthropic2024mcp} & 
\cmark & \xmark & \xmark & \xmark & \xmark & \xmark \\

Google A2A~\cite{google2025a2a} & 
\cmark & \xmark & \xmark & \xmark & \xmark & \xmark \\

Microsoft NLWeb~\cite{microsoft2025nlweb} & 
\cmark & \xmark & \xmark & \xmark & \xmark & \xmark \\

W3C Agent Web CG~\cite{w3c2025agentwebcg} & 
\cmark & $\sim$ & \xmark & \xmark & \xmark & \xmark \\

Cloudflare Pay-per-crawl~\cite{cloudflare2025block} & 
$\sim$ & \xmark & $\sim$ & \xmark & \xmark & \xmark \\

Nguyen et al.~\cite{nguyen2025agenteconomy} & 
\xmark & \xmark & \cmark & \xmark & \xmark & \xmark \\

Shumailov et al.~\cite{shumailov2023curse} & 
\xmark & \xmark & \xmark & \xmark & \cmark & \xmark \\

C2PA~\cite{c2pa2024} & 
\xmark & \xmark & \xmark & \xmark & \xmark & $\sim$ \\

Kirchenbauer et al.~\cite{kirchenbauer2023watermark} & 
\xmark & \xmark & \xmark & \xmark & $\sim$ & $\sim$ \\

Bender et al.~\cite{bender2021stochastic} & 
\xmark & \xmark & \xmark & \xmark & $\sim$ & \xmark \\

Evans~\cite{evans2008economics} & 
\xmark & \xmark & \cmark & \xmark & \xmark & \xmark \\

Wu~\cite{wu2016attention} & 
\xmark & \xmark & \cmark & \xmark & \xmark & \xmark \\

\midrule
\textbf{This paper (proposed)} & 
\cmark & \cmark & \cmark & \cmark & \cmark & \cmark \\
\bottomrule
\end{tabular}
\end{adjustbox}
\end{table*}

\section{The Human-Centric Web: A Three-Layer Diagnosis}
\label{sec:diagnosis}

The World Wide Web was not designed with a single deliberate 
choice to exclude non-human participants — rather, its 
human-centricity emerged organically from three decades of 
architectural decisions, each reasonable in isolation, that 
collectively produce a web hostile to agent-first interaction. 
In this section, we diagnose this incompatibility systematically 
across three layers: access, economics, and content. For each 
layer, we identify the foundational human-centric assumption, 
demonstrate empirically how that assumption fails under agent 
interaction, and characterize the nature of the resulting 
incompatibility. Proposed solutions to each layer are deferred 
to Sections~\ref{sec:diagnosis:access}, \ref{sec:diagnosis:economics}, and 
\ref{sec:diagnosis:content} respectively.

\subsection{Layer 1: The Access Model}
\label{sec:diagnosis:access}

\subsubsection{The Human-Centric Assumption}

The web's access model was designed around the presumption of 
human visitors. When a human navigates to a website, the 
interaction carries a set of implicit guarantees that the 
web's architecture has never needed to make explicit: the 
visitor is a single individual, operating at human speed, 
consuming content for personal use, and subject to social 
and legal accountability for their actions. These implicit 
guarantees justified the web's foundational philosophy of 
open access — articulated by Berners-Lee as the principle 
that any client should be able to retrieve any resource 
without prior negotiation \cite{berners1989information}. 
Security mechanisms such as CAPTCHAs operationalize this 
assumption directly, treating human biological capability 
as the access credential \cite{von2003captcha}.

The only formal mechanism for non-human access management 
is the \texttt{robots.txt} standard, introduced in 1994 
\cite{koster1994robots}. Designed for search engine crawlers 
operating under an implicit social contract — crawl freely, 
return traffic — \texttt{robots.txt} is an honor system with 
no enforcement mechanism, no concept of agent identity, no 
support for graduated access levels, and no capacity to 
distinguish between a personal assistant agent acting for 
one user and a mass scraper extracting content for commercial 
training \cite{cloudflare2025block}. It was adequate when 
the only non-human visitors were well-behaved search crawlers. 
It is wholly inadequate for the agent-first web.

\subsubsection{Empirical Evidence of Failure}

The incompatibility of the current access model with agent 
interaction is now empirically documented at infrastructure 
scale. Cloudflare, which routes over 16\% of global internet 
traffic, moved in July 2025 to block AI crawlers by default 
— the first time in the web's history that a major 
infrastructure provider has treated non-human access as an 
opt-in rather than a default right \cite{cloudflare2025block}. 
Between July 2025 and January 2026, the number of websites 
actively blocking AI crawlers was nearly seven times the 
number blocking traditional search crawlers such as Googlebot 
\cite{cloudflare2026google, astride}. Raw requests from GPTBot grew 
147\% from July 2024 to July 2025, while Meta-ExternalAgent 
grew 843\% over the same period \cite{cloudflare2025bots}, 
demonstrating that the scale asymmetry between human and 
agent access is not marginal but orders of magnitude.

The economic asymmetry driving this blocking response is 
equally stark. Cloudflare data from mid-2025 reveals that 
Google's search crawler — operating under the traditional 
crawl-for-traffic social contract — crawled approximately 
14 pages per human referral generated. By contrast, 
OpenAI's crawler exhibited a crawl-to-referral ratio of 
1,700:1, and Anthropic's crawler 73,000:1 
\cite{cloudflare2025ratio}. From a publisher's perspective, 
this ratio represents pure extraction: content consumed at 
massive scale with essentially zero value returned. The 
blocking response is rational given this asymmetry — but 
it is a symptom of architectural failure, not a solution.

\subsubsection{The Nature of the Incompatibility}

The fundamental incompatibility at the access layer is not 
that agents access content — humans do that too — but that 
the web has no mechanism to distinguish between access 
types that carry very different implications:

\begin{itemize}
    \item A \textit{personal agent} acting for one user, 
    consuming content on their behalf as a proxy for a 
    human visit — analogous to a human reading a webpage.

    \item A \textit{search agent} crawling content to build 
    an index that returns traffic to publishers — analogous 
    to Googlebot under the original social contract.

    \item A \textit{training crawler} bulk-extracting content 
    to train commercial AI models — a use case with no human 
    analog and genuinely novel economic implications.

    \item A \textit{malicious bot} probing for 
    vulnerabilities, scraping for competitive intelligence, 
    or performing denial-of-service — which should be blocked 
    regardless of whether the visitor is human or agent.
\end{itemize}

Because current infrastructure cannot distinguish these 
cases, the default response is to block all non-human 
access indiscriminately — effectively treating a personal 
assistant agent acting for a paying subscriber with the 
same hostility as a malicious scraper. This is the access 
layer's core failure: the absence of an agent identification 
and intent declaration mechanism that would enable graduated, 
context-appropriate responses rather than binary 
block-or-allow decisions.

\subsection{Layer 2: The Economic Model}
\label{sec:diagnosis:economics}

\subsubsection{The Human-Centric Assumption}

The web's economic model is built on the attention economy 
— a two-sided market in which publishers provide content 
to attract human attention, and advertisers pay to access 
that attention \cite{evans2008economics, wu2016attention}. 
Every economic mechanism the web has developed — advertising 
impressions, click-through rates, pageviews, time-on-site, 
subscription paywalls — is a proxy for human attention. The 
implicit assumption is that value flows when a human engages 
with content: seeing an advertisement, clicking a link, 
spending time reading. This model sustained a diverse 
publishing ecosystem for three decades, funding journalism, 
research, creative work, and open knowledge resources.

The attention economy's dependence on human engagement as 
its fundamental unit of value is not incidental — it is 
structural. Remove the human from the content consumption 
loop and the entire economic architecture collapses, because 
there is no attention to monetize, no click to count, and 
no impression to sell.

\subsubsection{Empirical Evidence of Failure}

AI agents are removing humans from the content consumption 
loop at measurable and accelerating scale. Zero-click 
searches — where AI systems synthesize answers without 
routing users to source content — now account for 
approximately 60\% of all Google queries 
\cite{semrush2025zeroclicks}. In AI-native search modes, 
this figure rises to 93\% \cite{semrush2025zeroclicks}. 
Only 1\% of users click links inside an AI Overview 
\cite{pew2025aioverview}. Click-through rates at position 
one in search results have fallen from 27\% to 11\% as AI 
Overviews have become pervasive \cite{sistrix2026ctr}. The 
downstream impact on publishers is severe and accelerating 
— HubSpot reported organic traffic declines of 70--80\%, 
Chegg lost 49\% of its traffic, and NPR characterized the 
shift as an extinction-level event for online news 
\cite{sistrix2026ctr}.

\begin{figure}[H]
\centering
\includegraphics[width=\textwidth]{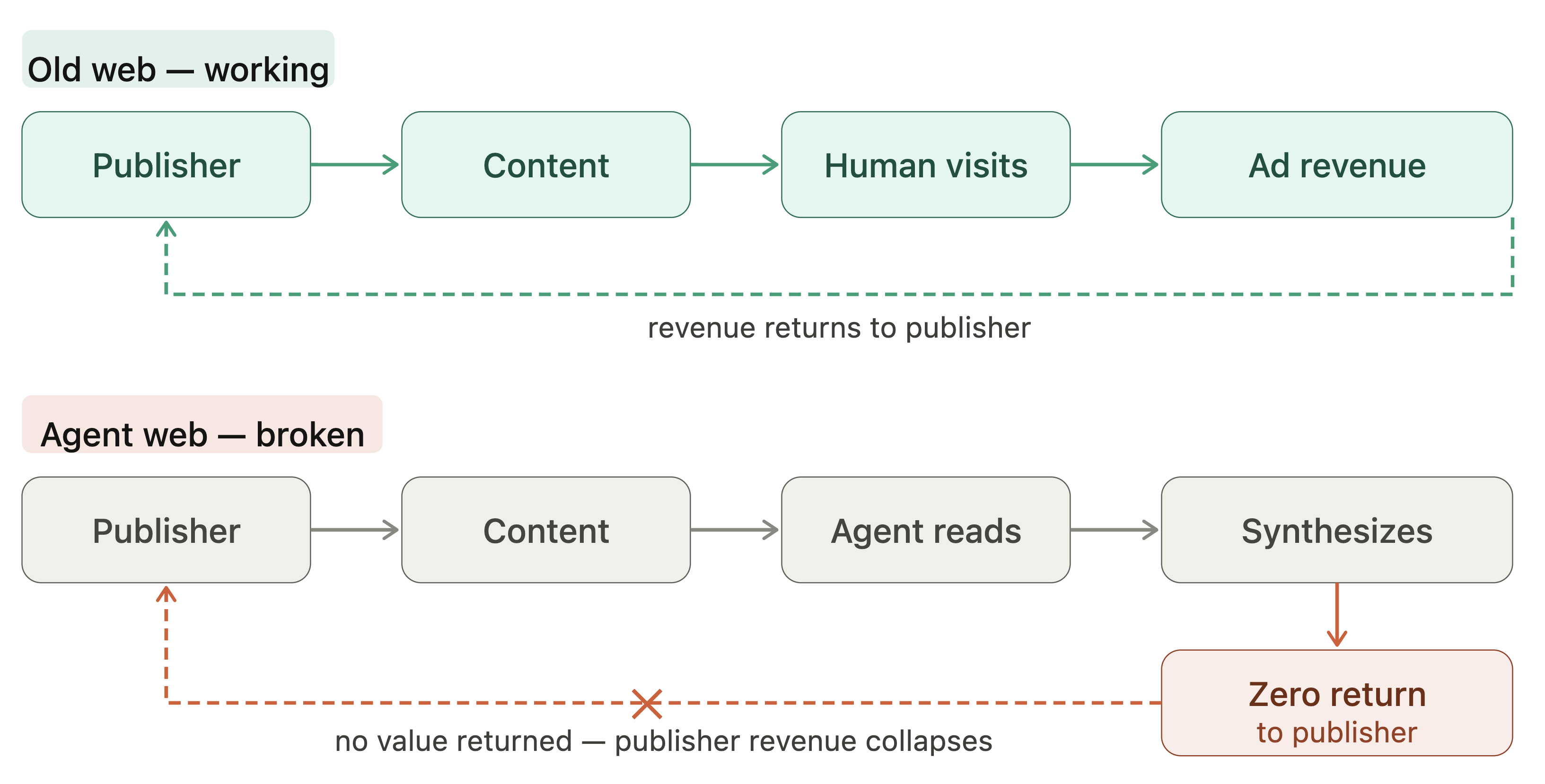}
\caption{The broken economic loop. In the human web (top), 
content consumption generates advertising revenue that 
returns to publishers, sustaining the content ecosystem. 
In the agent web (bottom), agents retrieve and synthesize 
content on behalf of users without generating pageviews, 
clicks, or impressions — delivering value to users while 
returning zero economic signal to publishers. This 
structural asymmetry is the primary driver of blanket 
agent blocking.}
\label{fig:economic}
\end{figure}

The economic asymmetry driving this blocking response is 
equally stark. Cloudflare data from mid-2025 reveals that 
Google's search crawler — operating under the traditional 
crawl-for-traffic social contract — crawled approximately 
14 pages per human referral generated. By contrast, 
OpenAI's crawler exhibited a crawl-to-referral ratio of 
1,700:1, and Anthropic's crawler 73,000:1 
\cite{cloudflare2025ratio}. Zero-click searches now account 
for approximately 60\% of all Google queries 
\cite{semrush2025zeroclicks}, rising to 93\% in AI-native 
search modes, with click-through rates at position one 
falling from 27\% to 11\% \cite{sistrix2026ctr}. 
Figure~\ref{fig:economic} illustrates how the human web's 
working value chain — where content consumption generates 
ad revenue that returns to publishers — is structurally 
broken under agent interaction, where content is consumed 
at scale with zero economic return.

\subsubsection{The Nature of the Incompatibility}

The economic incompatibility has three distinct components 
that must be addressed separately:

\textbf{The consumption problem.} When an agent retrieves 
and synthesizes content on behalf of a user, the publisher 
receives no economic signal — no impression, no click, no 
pageview. The value delivered to the user is real; the 
value returned to the publisher is zero. This is not a 
problem of agent malice but of architectural mismatch: the 
web has no mechanism for value exchange at the point of 
agent consumption.

\textbf{The attribution problem.} When an agent synthesizes 
an answer from multiple sources, there is no mechanism for 
fractional attribution — no way to record that source A 
contributed 40\% of the answer's informational content and 
source B contributed 60\%, and to distribute economic value 
accordingly. Human-web economics were never required to 
solve this problem because humans visit one page at a time.

\textbf{The production problem.} As AI generates increasing 
proportions of web content, the human creative labor that 
justified content economics — the journalist's 
investigation, the researcher's analysis, the expert's 
judgment — is displaced. An economic model that does not 
distinguish between human-authored and AI-generated content 
provides no incentive for the human contribution that 
grounds content quality and epistemic integrity.

\subsection{Layer 3: The Content Model}
\label{sec:diagnosis:content}

\subsubsection{The Human-Centric Assumption}

The web's content architecture was designed for human 
perception. HTML encodes visual layout, typography, color, 
and interactive behavior — all properties relevant to human 
reading experience and irrelevant to machine comprehension 
\cite{raggett1999html}. Search engine optimization, link 
structures, and content discovery mechanisms were designed 
around human attention patterns and search intent 
\cite{brin1998anatomy}. Even the semantic web effort — 
intended to make content machine-readable — was designed 
to augment human-targeted HTML rather than replace it 
\cite{berners2001semantic}.

The result is a content layer that is simultaneously 
over-specified for human rendering — carrying vast amounts 
of visual layout information irrelevant to agents — and 
under-specified for machine comprehension, lacking the 
semantic structure, provenance information, and intent 
metadata that agents need to evaluate and use content 
reliably.

\subsubsection{Empirical Evidence of Failure}

The inefficiency of current web content formats for agent 
consumption is quantifiable. \cite{webmcp2025, towards-rai-xai} demonstrated 
that agent-optimized content delivery — stripping visual 
rendering overhead and providing structured semantic content 
— achieves a 67.6\% reduction in token usage compared to 
standard HTML delivery. Since token consumption is directly 
proportional to computational cost in LLM-based agents, 
this represents a massive structural inefficiency imposed 
on every agent interaction with the current web.

\begin{figure}[H]
\centering
\includegraphics[width=\textwidth]{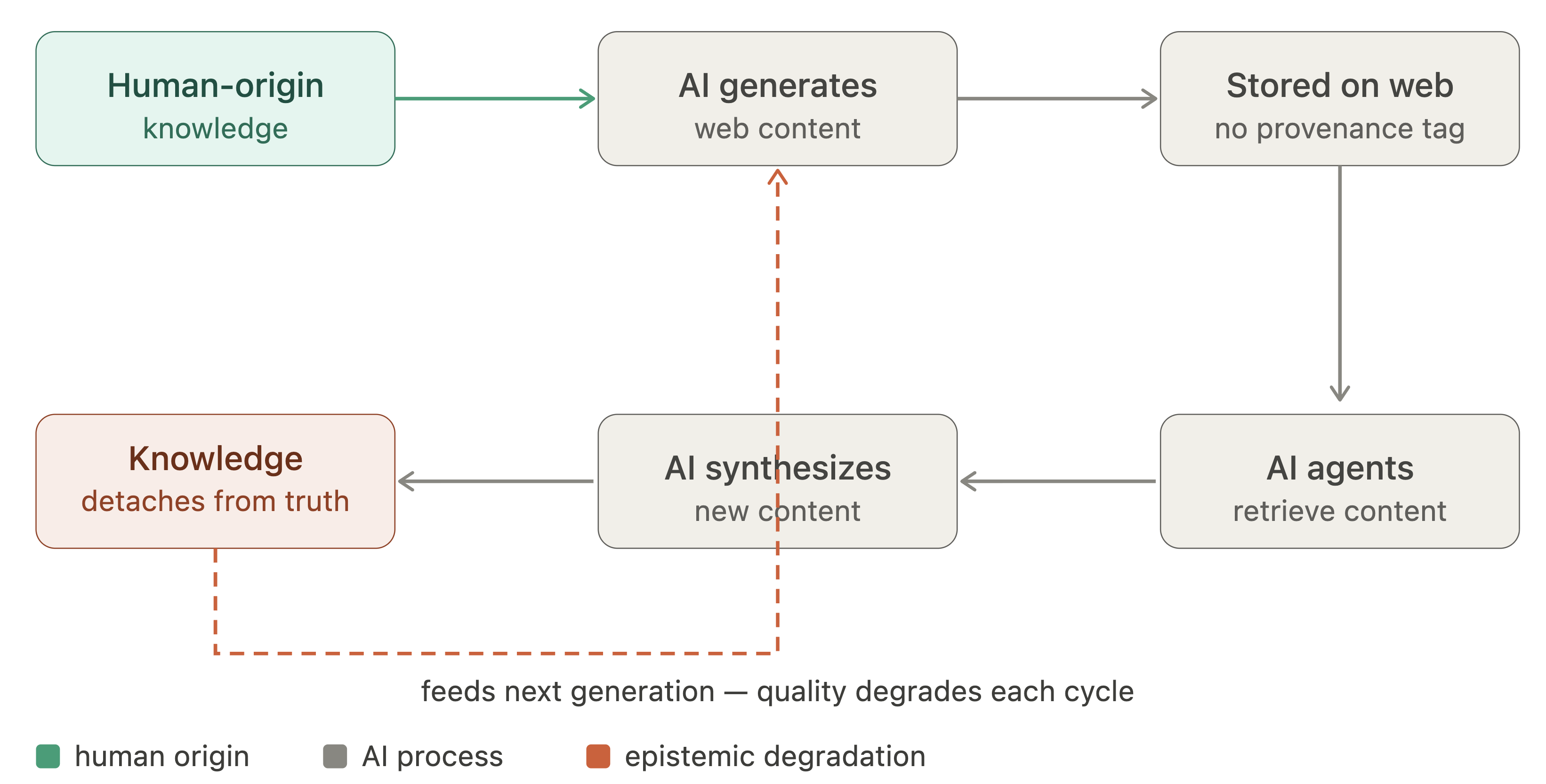}
\caption{The epistemic recursion loop. Human-origin 
knowledge enters the web through AI-generated content. 
AI agents retrieve and synthesize this content to produce 
further content, which re-enters the web for subsequent 
retrieval. Each cycle amplifies statistical artifacts and 
biases while reducing diversity, progressively detaching 
web knowledge from human ground truth. The dashed arrow 
represents the feedback path that sustains the loop across 
generations. Unlike model collapse \cite{shumailov2023curse}, 
which describes degradation within a single training 
pipeline, epistemic recursion describes a web-scale 
phenomenon operating across the entire AI content 
ecosystem.}
\label{fig:epistemic}
\end{figure}

Beyond format inefficiency, the content layer faces a 
structural integrity problem with no precedent in the human 
web. \cite{shumailov2023curse} demonstrated the phenomenon 
of model collapse — the progressive degradation of AI model 
quality when trained iteratively on AI-generated data. Each 
generation of AI content introduces statistical artifacts, 
amplifies existing biases, and loses the diversity that 
characterizes human-origin content. When this content 
re-enters the training pipeline, these artifacts compound. 
The web, as the primary reservoir of training data for 
large language models, is thus at risk of becoming a source 
of progressively degraded knowledge — not through any 
deliberate action but through the structural dynamics of AI 
content generation at scale. We term this phenomenon 
\textit{epistemic recursion}: the self-referential loop in 
which AI-generated content becomes the input for future AI 
content generation, progressively detaching web knowledge 
from human ground truth. Figure~\ref{fig:epistemic} 
illustrates this loop and its compounding degradation 
across generations.

\subsubsection{The Nature of the Incompatibility}

We identify three distinct content-layer incompatibilities:

\textbf{The format incompatibility.} Current web content 
formats — HTML, JavaScript-rendered pages, PDF documents 
— carry significant visual rendering overhead that is 
meaningless to agents and computationally expensive to 
process. Agents require structured semantic content: clean 
text, explicit metadata, machine-readable relationships 
between concepts. The web currently has no standard format 
serving this requirement, though recent proposals such as 
\texttt{llms.txt} \cite{llmstxt2024} represent early steps 
toward agent-readable content declarations.

\textbf{The provenance incompatibility.} Current web 
content carries no standard mechanism for declaring its 
origin, derivation chain, or the degree of human oversight 
involved in its production. An agent retrieving a web page 
cannot determine whether it is reading a human expert's 
peer-reviewed analysis, an AI-generated content farm 
article, or a human-AI collaborative piece. This absence 
of provenance metadata prevents agents from making 
quality-aware content decisions and enables the epistemic 
recursion problem described above.

\textbf{The discoverability incompatibility.} Search 
engine optimization was built to attract human attention 
through search rankings optimized for human query 
patterns. There is no equivalent agent discoverability 
standard — no mechanism by which an agent can determine 
which site holds authoritative data on a topic, what 
content formats a site supports, what access terms apply, 
or what semantic capabilities a site exposes. Agents 
currently navigate content discovery through human-facing 
search interfaces, producing significant inefficiency and 
unreliability.

\subsection{Summary: Three Layers, One Failure}
\label{sec:diagnosis:summary}

\begin{figure}[H]
\centering
\includegraphics[width=\textwidth]{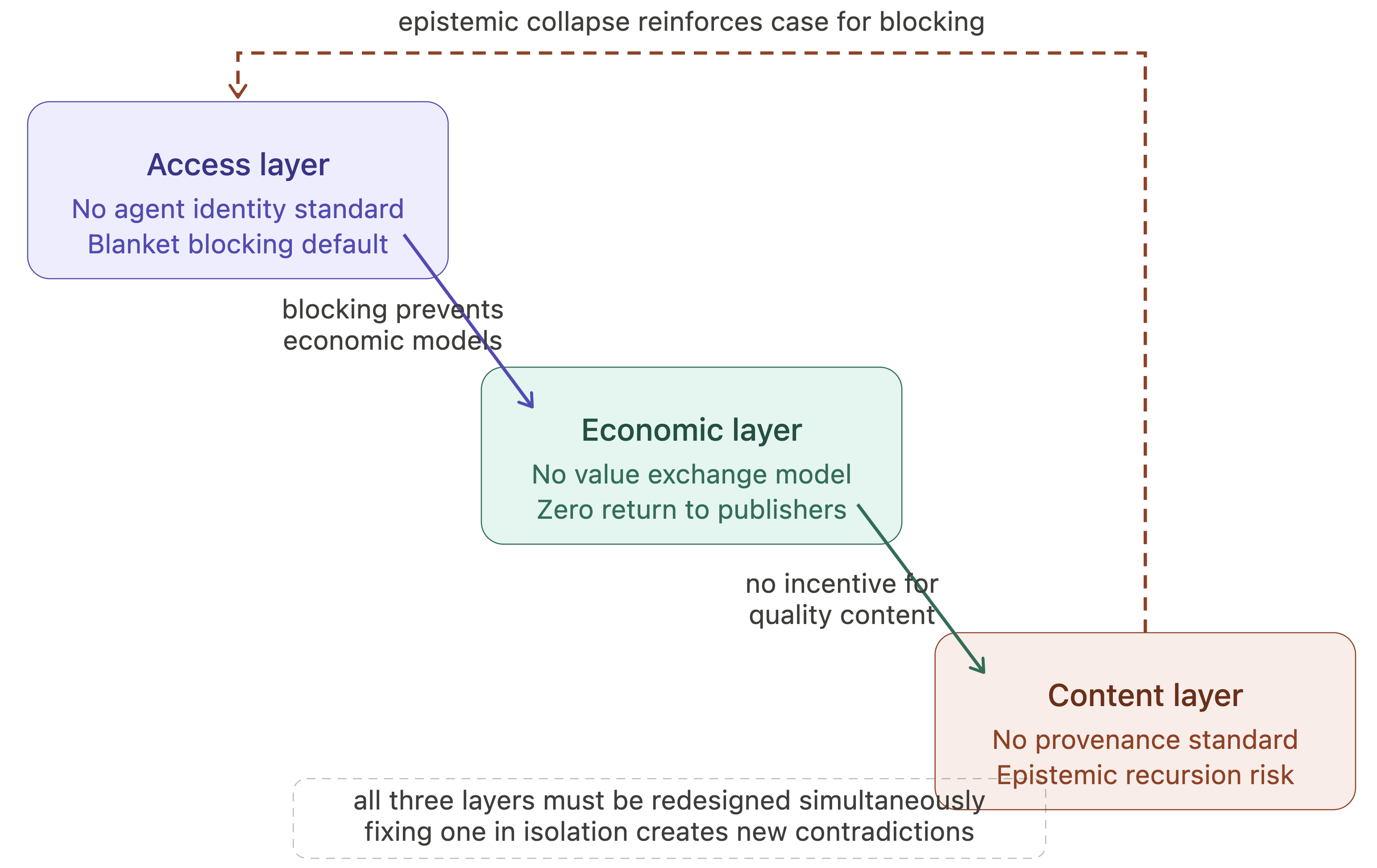}
\caption{Three-layer failure coupling. The access, economic, 
and content layer failures of the human-centric web are 
deeply interdependent. Blanket blocking (access) prevents 
the development of economic models; absent economic models 
eliminate incentives for quality content production; 
degraded content quality reinforces the case for blocking. 
The dashed arc represents the feedback path completing the 
cycle. Fixing any single layer in isolation creates new 
contradictions — all three must be redesigned 
simultaneously.}
\label{fig:layers}
\end{figure}

The three failures diagnosed in this section are not 
independent — they are deeply coupled in a self-reinforcing 
cycle. The access layer's inability to identify agent intent 
drives blanket blocking, which prevents the development of 
economic models for legitimate agent access. The absence of 
economic models eliminates the incentive for publishers to 
invest in high-quality, human-supervised content production. 
The collapse of human-supervised content production 
accelerates epistemic recursion, further degrading the 
web's knowledge quality and reinforcing the perception that 
agent access is extractive rather than valuable — which in 
turn strengthens the case for blanket blocking. 
Figure~\ref{fig:layers} illustrates this interdependency. 
Addressing any single layer in isolation — as existing work 
has largely done — produces solutions that are undermined 
by the failures of the other two. A coherent redesign must 
address all three layers simultaneously, grounded in a 
unified philosophical foundation, which we develop in the 
following sections.

Table~\ref{tab:diagnosis} summarizes the three-layer 
diagnosis across the key dimensions of assumption, 
empirical evidence, and nature of incompatibility.

\begin{table}[H]
\centering
\caption{Three-layer diagnosis of human-centric web 
failure under agent interaction.}
\label{tab:diagnosis}
\begin{adjustbox}{width=1\textwidth}
\begin{tabular}{lccc}
\toprule
\thead{Layer} & 
\thead{Human-centric\\assumption} & 
\thead{Empirical\\evidence} &
\thead{Nature of\\incompatibility} \\
\midrule
\textbf{Access} & 
\makecell{Visitors are single humans \\at human speed} & 
\makecell{7$\times$ more sites \\blocking AI vs Googlebot} \cite{cloudflare2026google} &
\makecell{No mechanism to distinguish \\ agent types or intent} \\
\addlinespace
\textbf{Economic} & 
\makecell{Value flows through human \\ attention and clicks} & 
\makecell{73,000:1 crawl-to-referral ratio \cite{cloudflare2025ratio}; \\ 93\% zero-click in AI search} \cite{semrush2025zeroclicks} &
\makecell{No value exchange at point of agent \\ consumption; attribution impossible} \\
\addlinespace
\textbf{Content} & 
\makecell{Content is human-authored and \\ rendered for human perception} & 
\makecell{67.6\% token overhead in HTML \cite{webmcp2025}; \\ model collapse under AI training \cite{shumailov2023curse}} &
\makecell{Format overhead; no provenance standard; \\ epistemic recursion risk} \\
\bottomrule
\end{tabular}
\end{adjustbox}
\end{table}

\section{The Access Layer Redesign}
\label{sec:access}

Having diagnosed the three-layer failure of the human-centric 
web in Section~\ref{sec:diagnosis}, we now turn to proposed 
solutions. This section addresses the access layer, proposing 
a principled redesign grounded in a single philosophical 
anchor: that AI agents acting on behalf of humans are 
first-class web citizens entitled to the same presumption 
of access as the humans they represent.

\subsection{The Agent-as-Human-Proxy Principle}
\label{sec:access:proxy}

The web's founding access philosophy — articulated by 
Berners-Lee as the principle that any client should be 
able to retrieve any resource without prior negotiation 
\cite{berners1989information} — was grounded in a 
presumption of human visitors. We propose extending this 
presumption to AI agents through what we term the 
\textit{agent-as-human-proxy principle}:

\begin{quote}
\textit{An AI agent acting on behalf of a human user 
should inherit the same presumption of access as that 
human — no more, no less. The agent's access rights, 
obligations, and economic responsibilities are 
determined by the human it represents, not by its 
nature as an automated system.}
\end{quote}

This principle has several important implications. First, 
it establishes that the relevant unit of access is the 
human-agent pair, not the agent in isolation. An anonymous 
human can browse freely — their agent should too. A 
subscribed human has paid for access — their agent should 
inherit it. A human cannot legally bulk-scrape a million 
pages for commercial resale — neither can their agent. 
Second, it implies that blanket blocking of agents is 
philosophically unjustifiable when those agents represent 
humans who would otherwise be welcome. Third, it identifies 
the two genuinely novel cases that fall outside the 
human-proxy principle — commercial training use and 
multi-user content aggregation — as the only cases 
requiring new access rules rather than extensions of 
existing human-web norms.

The agent-as-human-proxy principle does not eliminate 
the need for agent identification — it defines what that 
identification must accomplish. Rather than requiring 
agents to prove full identity (which raises privacy 
concerns and creates friction), agents need only declare 
the \textit{behavioral context} that makes the proxy 
relationship legible: who they represent, at what scale, 
and for what purpose.

\subsection{Agent Identification Metadata}
\label{sec:access:metadata}

The central technical contribution of this section is 
the proposal for \textit{agent identification metadata} 
— a set of standardized HTTP request headers that allow 
agents to declare their identity, represented user, and 
intent to web servers. This proposal is grounded in a 
direct analogy to existing browser identification 
mechanisms.

When a browser makes an HTTP request, it sends a set 
of headers that identify its capabilities and context 
\cite{fielding1999http}:

\begin{verbatim}
User-Agent: Mozilla/5.0 (Macintosh; Intel Mac OS X)
Accept: text/html,application/xhtml+xml
Accept-Language: en-US,en;q=0.9
\end{verbatim}

Servers read these headers to serve appropriate content 
— mobile vs desktop layouts, supported media types, 
preferred language versions. No new protocol is required; 
the mechanism is embedded in the existing HTTP 
infrastructure that every web server already supports.

We propose an analogous mechanism for agents:

\begin{verbatim}
Agent-Identity: claude/sonnet-4
Agent-Represents: user/anonymous
Agent-Intent: personal-use
Agent-Auth: Bearer <delegation-token>
Agent-Rate-Class: free
\end{verbatim}

The semantics of each header are as follows. 
\texttt{Agent-Identity} declares the agent system and 
version — analogous to \texttt{User-Agent} for browsers. 
\texttt{Agent-Represents} declares the relationship 
between the agent and the human it serves — anonymous, 
authenticated, or subscribed. \texttt{Agent-Intent} 
declares the purpose of the request from a controlled 
vocabulary: \texttt{personal-use}, \texttt{search}, 
\texttt{training}, \texttt{commercial}, or 
\texttt{research}. \texttt{Agent-Auth} carries a 
delegation token linking the agent to a human user's 
existing credentials — enabling subscription 
inheritance without exposing personal identity. 
\texttt{Agent-Rate-Class} declares the agent's 
expected consumption tier, enabling servers to 
pre-authorize appropriate rate limits.

\begin{figure}[H]
\centering
\includegraphics[width=\textwidth]{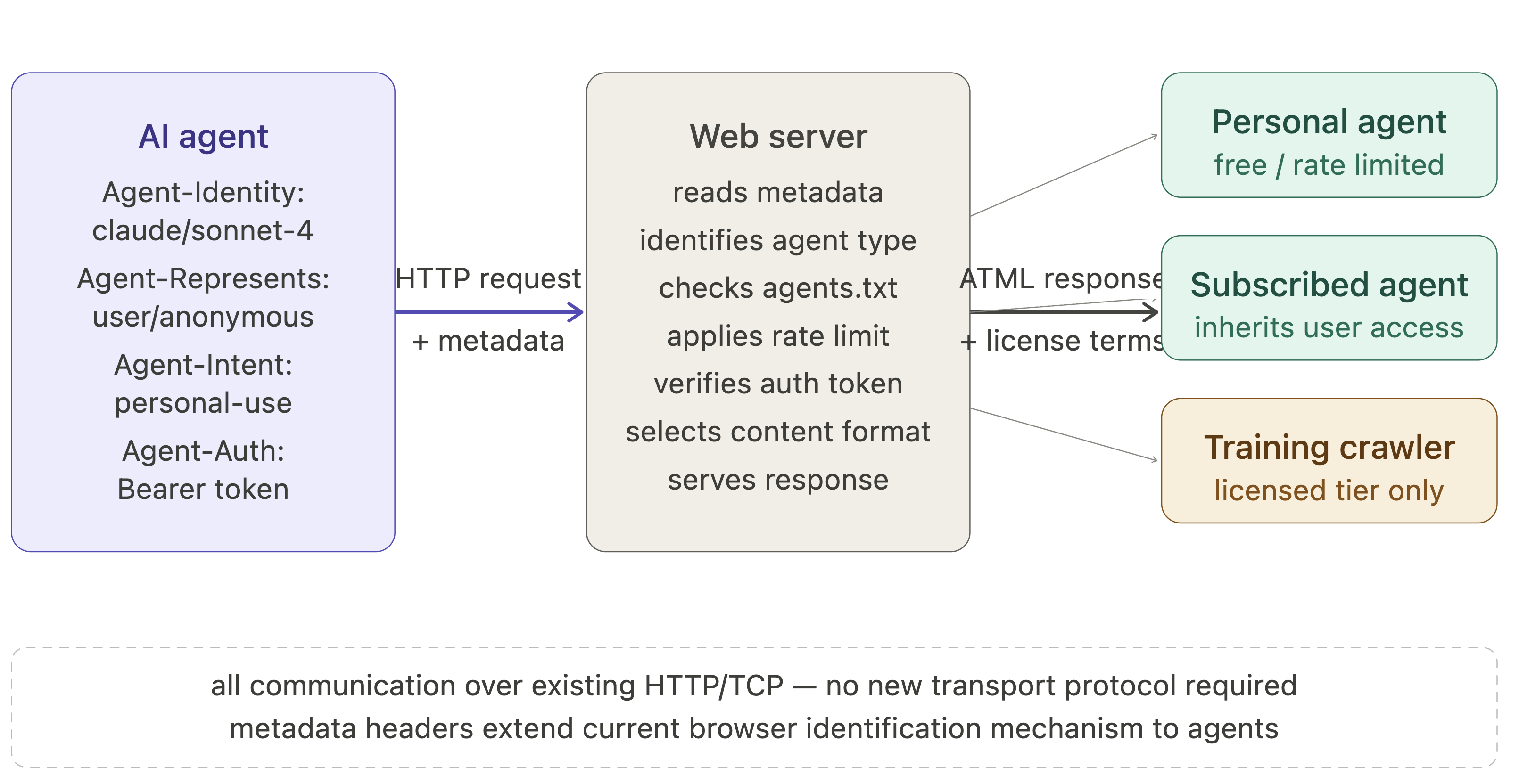}
\caption{Agent identification metadata flow. An AI agent 
sends standardized metadata headers alongside its HTTP 
request, declaring identity, represented user, and 
intent. The web server reads these headers to identify 
the agent type, verify authorization, apply appropriate 
rate limits, and select the correct content format 
(ATML for agents, HTML for humans). All communication 
occurs over existing HTTP/TCP infrastructure — no new 
transport protocol is required. The metadata mechanism 
directly extends the browser User-Agent identification 
pattern already ubiquitous in web infrastructure.}
\label{fig:metadata}
\end{figure}

Figure~\ref{fig:metadata} illustrates the full 
request-response flow enabled by agent identification 
metadata, showing how servers can distinguish agent 
types and respond with graduated access rather than 
binary blocking.

This proposal requires no new transport protocol — 
everything operates over existing HTTP infrastructure 
that every web server already supports. It is backward 
compatible: servers that do not recognize agent metadata 
headers simply ignore them, serving their existing 
content as before. Adoption can be incremental, with 
agent-aware servers gaining the ability to serve 
optimized content and apply graduated access controls 
while non-aware servers continue to function normally.

A critical security consideration is metadata 
verification. Unlike browser User-Agent headers, which 
carry no cryptographic guarantee, agent metadata headers 
should be cryptographically signed to prevent 
impersonation — the Perplexity case demonstrated that 
bad actors will modify agent identifiers to circumvent 
restrictions \cite{perplexity2024sharing}. We propose 
that the \texttt{Agent-Identity} header be accompanied 
by a cryptographic signature verifiable against a 
public key registry maintained by registered agent 
operators, analogous to DKIM signatures for email 
\cite{allman2007dkim}. This does not prevent all 
impersonation but raises the cost significantly above 
simple header modification.

\subsection{Rate Limiting as the New Default}
\label{sec:access:ratelimit}

The agent-as-human-proxy principle implies that blanket 
blocking of agents is not a legitimate long-term access 
policy — it is a reactive measure taken in the absence 
of better tools. Agent identification metadata provides 
those tools. Once servers can distinguish agent types 
and intents, the appropriate response is graduated rate 
limiting rather than binary blocking.

\begin{figure}[H]
\centering
\includegraphics[width=\textwidth]{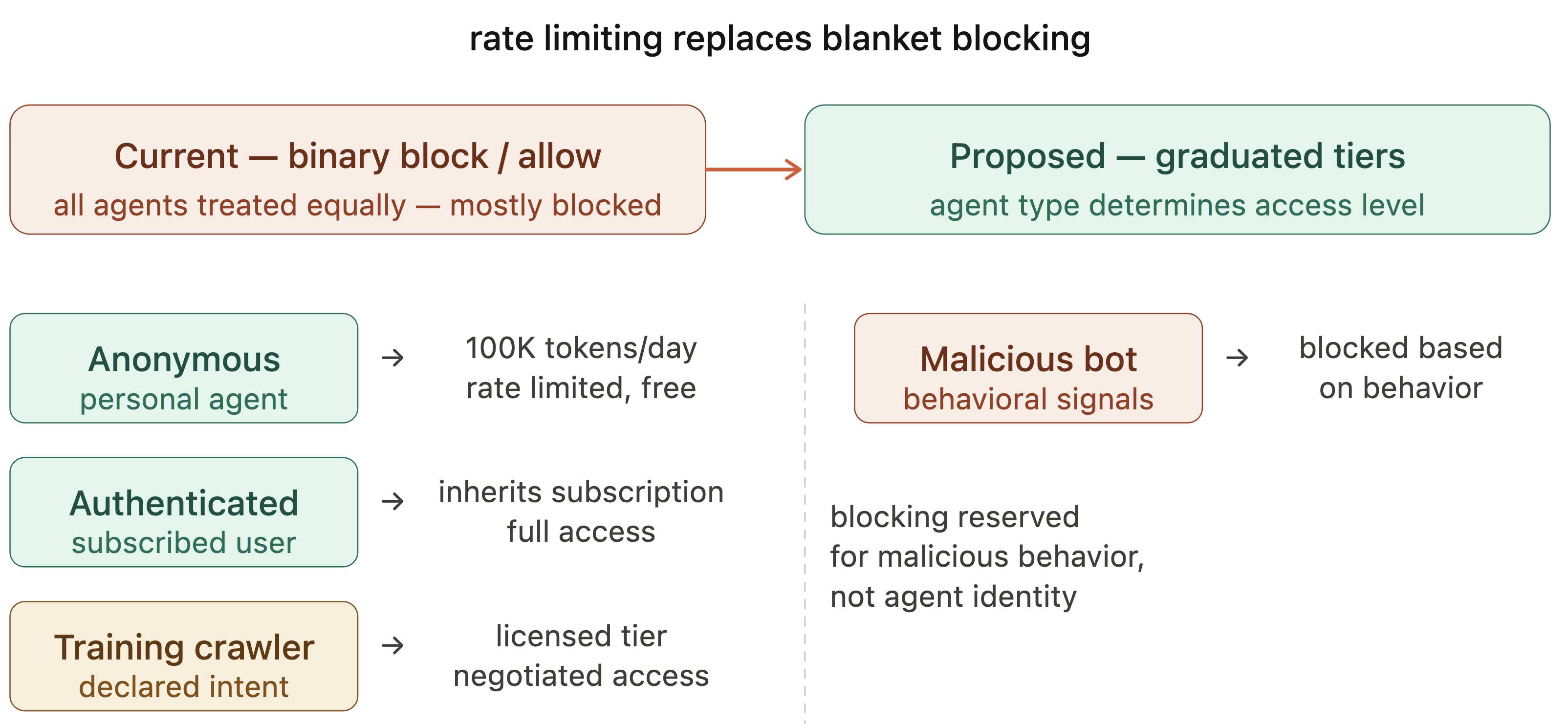}
\caption{Graduated rate limiting model replacing blanket 
blocking. Agent access is tiered by declared intent and 
authentication level rather than binary blocked or 
allowed. Anonymous personal agents receive rate-limited 
free access, equivalent to anonymous human browsing. 
Authenticated agents inherit their represented user's 
subscription. Training crawlers access a licensed tier 
with negotiated terms. Only malicious behavior — not 
agent identity — triggers blocking. This model 
preserves the web's open access philosophy while 
providing servers with meaningful differentiation 
between agent types.}
\label{fig:ratelimit}
\end{figure}

We propose that the default access control posture for 
agent requests shift from \textit{block unless permitted} 
to \textit{rate limit unless escalated}, mirroring the 
web's existing posture toward anonymous human visitors. 
Figure~\ref{fig:ratelimit} illustrates the proposed 
tier model.

The specific rate limits for each tier are not mandated 
by this framework — they are determined by individual 
publishers based on their content economics and 
infrastructure capacity. The framework mandates only 
the structure: that tiers exist, that they are 
determined by intent and authentication level rather 
than by agent identity alone, and that blocking is 
reserved for behaviorally malicious access rather than 
applied to all non-human requests indiscriminately.

\subsection{Subscription Inheritance Protocol}
\label{sec:access:subscription}

Perhaps the most immediately actionable proposal in 
this paper is subscription inheritance — the principle 
that a human user's existing paid subscriptions should 
extend to their AI agents without requiring separate 
agent subscriptions. This is a direct application of 
the agent-as-human-proxy principle: if a human has 
paid for access, their agent proxy should have it too.

Technically, this requires a lightweight extension to 
existing OAuth 2.0 delegation mechanisms 
\cite{hardt2012oauth}. A user authenticates with a 
content provider as normal, obtaining an access token. 
When authorizing their AI agent, the user generates a 
\textit{delegation token} — a derived credential that 
encodes:

\begin{itemize}
    \item The human user's subscription level and scope
    \item The agent system authorized to use it
    \item The permitted use cases (personal-use only, 
    no training, no redistribution)
    \item An expiry and revocation mechanism
\end{itemize}

The delegation token is passed as the 
\texttt{Agent-Auth} header in agent requests. The 
server verifies the token, confirms it maps to a valid 
subscription, and grants the agent the same content 
access the human subscriber would receive. The human 
user's subscription is not consumed or duplicated — 
it is extended to their proxy.

This mechanism has several advantages over alternative 
approaches. It requires no changes to existing 
subscription billing infrastructure. It preserves 
publisher revenue — the human still pays for 
subscription, the agent merely inherits it. It gives 
users meaningful control over what their agents can 
access on their behalf. And it is directly compatible 
with existing OAuth infrastructure deployed by virtually 
every major content platform.

\subsection{Dual-Layer Web Architecture}
\label{sec:access:duallayer}

The access layer redesign requires not only new 
identification and authorization mechanisms but a new 
approach to content delivery. We propose a 
\textit{dual-layer web architecture} in which the same 
domain serves both human-readable HTML and 
agent-optimized content — with the server selecting 
the appropriate format based on request metadata.

This is not a new concept in web architecture — 
content negotiation has been part of HTTP since its 
earliest versions \cite{fielding1999http}, enabling 
servers to serve different content types to different 
clients. The \texttt{Accept} header allows browsers 
to declare their preferred media types; servers 
respond with the best available match. We propose 
extending this mechanism to include an agent-optimized 
content type — which we call ATML (Agent Text Markup 
Language, developed in Section~\ref{sec:content}) 
— as a first-class content type alongside HTML.

A browser request declares:
\begin{verbatim}
Accept: text/html, application/xhtml+xml
\end{verbatim}

An agent request declares:
\begin{verbatim}
Accept: application/atml+xml, text/markdown
Agent-Identity: claude/sonnet-4
Agent-Intent: personal-use
\end{verbatim}

The server responds with HTML to the browser and ATML 
to the agent — the same content, differently structured. 
No separate domain, no separate publishing workflow. 
Publishers maintain one content repository; the server 
layer handles format negotiation.

\begin{figure}[H]
\centering
\includegraphics[width=\textwidth]{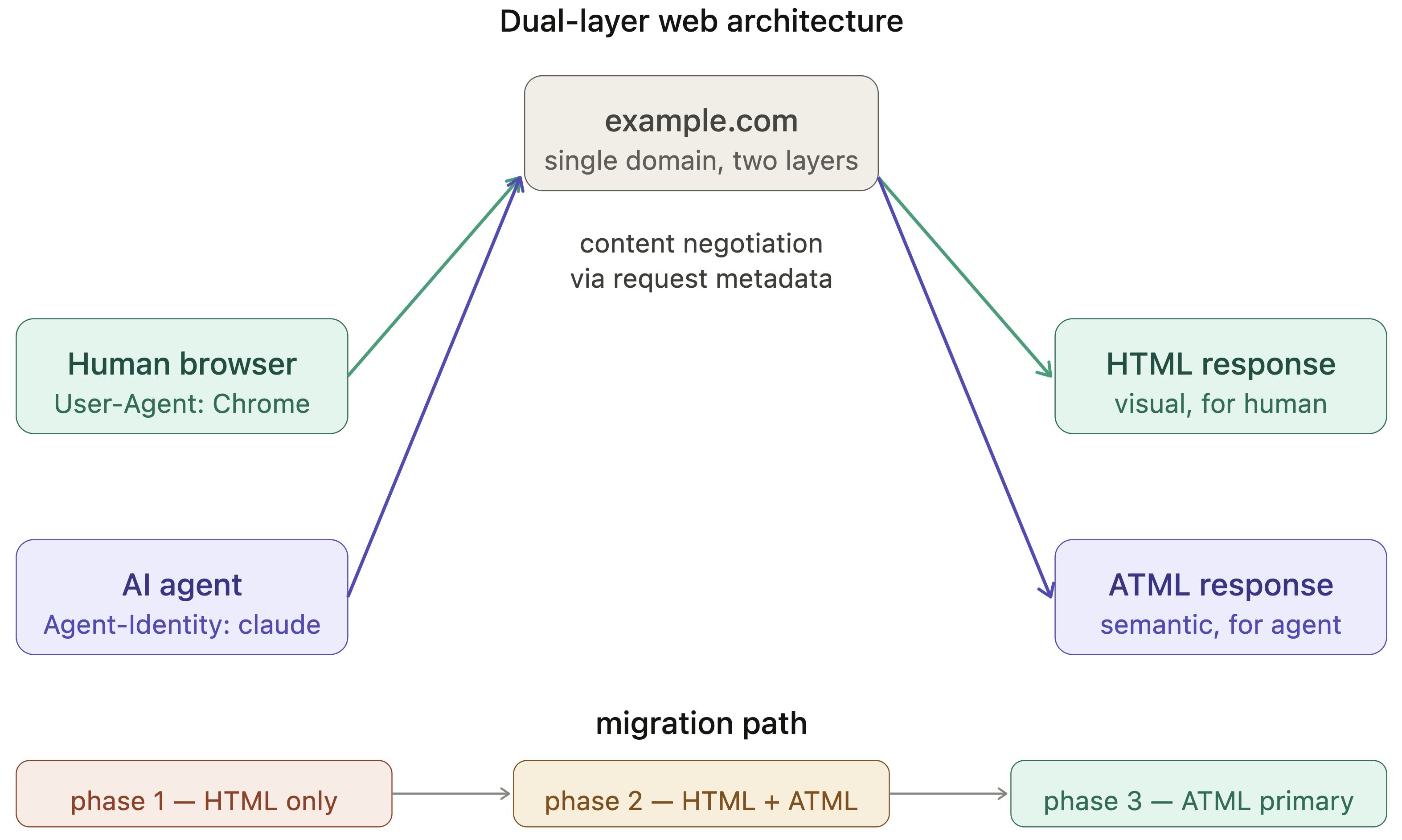}
\caption{Dual-layer web architecture and migration 
path. The same domain serves HTML responses to 
browser requests and ATML responses to agent 
requests, with format selection driven by request 
metadata. Phase 1 represents the current state 
(HTML only). Phase 2 introduces parallel ATML 
alongside HTML. Phase 3 establishes ATML as the 
primary format with HTML maintained for legacy 
compatibility — analogous to how HTTPS was added 
alongside HTTP before becoming the mandatory 
standard.}
\label{fig:duallayer}
\end{figure}

Figure~\ref{fig:duallayer} illustrates this 
architecture and the three-phase migration path from 
the current human-only web to an agent-first web.

The migration path is deliberately gradual. Phase 1 
represents the current state — human HTML only, 
with agent metadata headers added as a forward 
compatibility measure. Phase 2 introduces ATML 
as a parallel delivery format, with publishers 
adopting it incrementally. Phase 3 establishes 
ATML as the primary delivery format, with HTML 
maintained for legacy browser compatibility. This 
mirrors the HTTPS migration: years of 
coexistence, then a tipping point driven by 
security requirements, then universal adoption 
\cite{felt2017measuring}.

\subsection{A New agents.txt Standard}
\label{sec:access:agentstxt}

The final component of the access layer redesign is 
a replacement for \texttt{robots.txt} — a richer, 
machine-readable standard that declares a site's 
agent access policy in a way that the existing 
honor-system standard cannot. We propose 
\texttt{agents.txt}, a structured declaration file 
served at a well-known URL (\texttt{/.well-known/agents.txt}) 
that encodes:

\begin{verbatim}
# agents.txt example
[personal-agent]
allow: true
rate-limit: 100000 tokens/day
content-format: atml, markdown
auth-required: false

[subscribed-agent]
allow: true
rate-limit: unlimited
content-format: atml, markdown
auth-required: true
auth-type: delegation-token

[training-crawler]
allow: conditional
license-required: true
contact: licensing@publisher.com

[malicious-bot]
allow: false
\end{verbatim}

Unlike \texttt{robots.txt}, \texttt{agents.txt} is 
not an honor system — its declarations are enforced 
by the agent identification metadata mechanism. A 
server receiving a request with 
\texttt{Agent-Intent: training} can automatically 
apply the \texttt{[training-crawler]} policy declared 
in \texttt{agents.txt}, without manual configuration 
per crawler. Publishers gain granular control over 
agent access without resorting to blanket blocking. 
Agents gain clear, machine-readable signals about 
what access is available and under what terms — 
eliminating the current situation where agents 
must attempt access and interpret failures 
heuristically.

\subsection{Summary}
\label{sec:access:summary}

Table~\ref{tab:access} summarizes the access layer 
redesign proposals and their relationship to the 
failures diagnosed in Section~\ref{sec:diagnosis:access}.

\begin{table}[H]
\centering
\caption{Access layer redesign: diagnosed failures 
and proposed solutions.}
\label{tab:access}
\begin{adjustbox}{width=0.48\textwidth}
\begin{tabular}{p{3cm}p{4.5cm}}
\toprule
\thead{Diagnosed failure} & 
\thead{Proposed solution} \\
\midrule
No agent identity standard &
Agent identification metadata headers over HTTP \\
\addlinespace
Blanket blocking default &
Graduated rate limiting by intent and auth level \\
\addlinespace
No subscription extension &
OAuth delegation token for subscription inheritance \\
\addlinespace
HTML-only delivery &
Dual-layer architecture with ATML content negotiation \\
\addlinespace
robots.txt inadequacy &
agents.txt — rich machine-readable access policy \\
\bottomrule
\end{tabular}
\end{adjustbox}
\end{table}

Together these five proposals constitute a complete 
access layer redesign that preserves the web's 
founding philosophy of open access while providing 
the identification, authorization, and policy 
infrastructure necessary for servers to respond 
to agents appropriately rather than defensively. 
Critically, all proposals operate over existing 
HTTP/TCP infrastructure — no new transport protocol 
is required. The agent-first web is not a new 
internet; it is the existing internet with the 
missing metadata layer added.

\section{The Economic Layer Redesign}
\label{sec:economics}

The access layer redesign proposed in 
Section~\ref{sec:access} establishes how agents reach 
content. This section addresses what happens when they 
do — proposing an economic framework that replaces the 
broken attention economy with a principled value-exchange 
model. We argue that the economic redesign must be grounded 
in the same philosophical anchor as the access redesign: 
the agent-as-human-proxy principle. An agent's economic 
obligation mirrors that of the human it represents — 
no more, no less.

\subsection{The Attention Economy Cannot Be Patched}
\label{sec:economics:attention}

The web's attention economy was not designed — it 
emerged organically as advertising became the dominant 
funding mechanism for online content 
\cite{evans2008economics, wu2016attention}. Its core 
mechanism — human attention proxied by clicks and 
pageviews, monetized through advertising impressions 
— worked because it created a closed value loop: 
publishers produced content, humans consumed it, 
advertisers paid for access to that attention, and 
revenue returned to publishers who produced more 
content. Every layer of the web's architecture evolved 
to serve this loop: SEO to attract human attention, 
analytics to measure it, advertising networks to 
monetize it.

AI agents do not fit this loop. They do not see 
advertisements. They do not generate pageviews. They 
do not click. The value they deliver to users is real 
— task completion, information synthesis, decision 
support — but that value is entirely decoupled from 
the economic signals the attention economy was built 
to capture. This is not a marginal disruption. 
Zero-click searches now account for 93\% of queries 
in AI-native search modes \cite{semrush2025zeroclicks}, 
and click-through rates at position one have fallen 
from 27\% to 11\% \cite{sistrix2026ctr}. The attention 
economy is not being disrupted at the edges — it is 
collapsing at its foundation.

Reactive patches — Cloudflare's pay-per-crawl model, 
Perplexity's revenue sharing program 
\cite{perplexity2024sharing, cloudflare2025block} — 
address specific friction points without a principled 
framework. They treat agent access as a billing 
problem rather than a design problem. The result is 
an incoherent patchwork: some content is paywalled 
per crawl, some is blocked outright, some is freely 
scraped, and publishers have no consistent basis on 
which to make access decisions. A principled framework 
is needed.

\subsection{The Agent-as-Human-Proxy Economic Principle}
\label{sec:economics:proxy}

We propose that the economic redesign be grounded in 
a single principle that resolves the majority of agent 
content access cases without requiring new economic 
models:

\begin{quote}
\textit{An agent's economic obligation to a content 
publisher is equivalent to the economic obligation 
of the human it represents. If the content is free 
for anonymous humans, it is free for their agents. 
If the content requires a subscription, the agent 
inherits that subscription. The economic relationship 
is between the human and the publisher; the agent 
is a proxy, not a new economic entity.}
\end{quote}

This principle is powerful because it resolves most 
agent content access cases by reference to existing 
human-web economics — no new models required. It 
also identifies precisely where new models are needed: 
the genuinely novel cases for which no human analog 
exists. These are commercial training use — where 
an agent bulk-extracts content to train AI models, 
a use case no human performs — and multi-user 
aggregation — where a single agent serves content 
to millions of users simultaneously, a scale no 
human achieves. For these two cases, the 
human-proxy principle does not apply, and new 
economic models must be designed from first 
principles.

\subsection{The Intent-Based Economic Tier Model}
\label{sec:economics:tiers}

\begin{figure}[H]
\centering
\includegraphics[width=\textwidth]{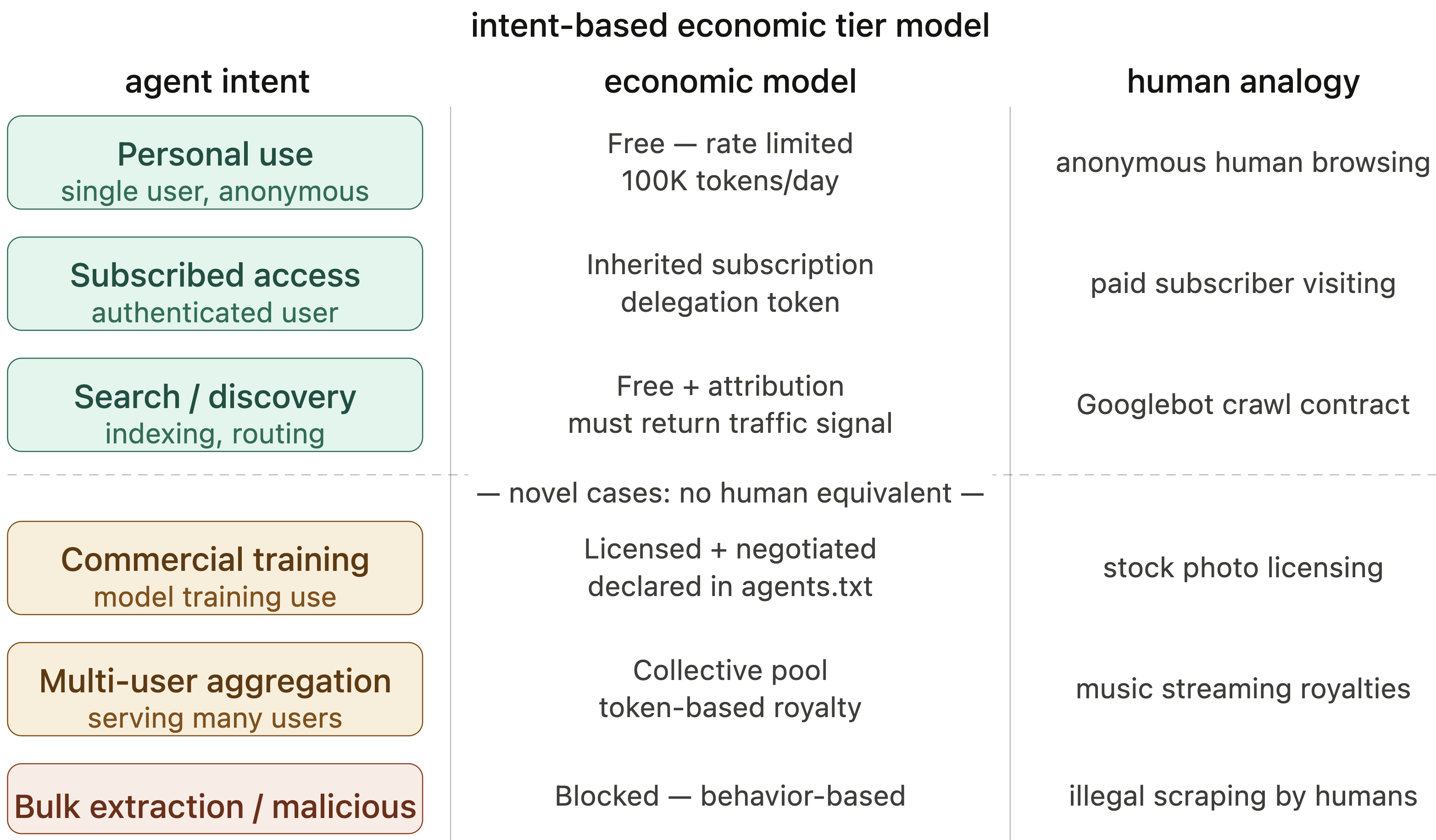}
\caption{Intent-based economic tier model. Agent 
content access behaviors are classified by declared 
intent and authentication level. The top three tiers 
— personal use, subscribed access, and search/discovery 
— are resolved by the agent-as-human-proxy principle, 
mapping directly to existing human-web economic models. 
The bottom three tiers — commercial training, 
multi-user aggregation, and bulk extraction — 
represent genuinely novel cases with no human analog, 
requiring new economic models. The dividing line 
between these groups is the boundary of the 
human-proxy principle.}
\label{fig:econotiers}
\end{figure}

Applying the agent-as-human-proxy principle produces 
a natural taxonomy of agent content access behaviors, 
each with an appropriate economic model. We term this 
the \textit{intent-based economic tier model}. 
Figure~\ref{fig:econotiers} presents the full model.

For the human-proxy tiers, the economic model is 
straightforward. A personal agent accessing free 
content on behalf of an anonymous user pays nothing 
— exactly as the user would. A subscribed agent 
accessing paywalled content inherits the user's 
subscription via the delegation token mechanism 
described in Section~\ref{sec:access:subscription}. 
A search agent crawling content for indexing operates 
under the traditional crawl-for-traffic social 
contract, with a machine-readable attribution 
requirement replacing the implicit honor system 
that \texttt{robots.txt} currently relies on.

For the novel tiers, new models are required. 
Commercial training access requires a licensing 
framework analogous to stock photo or music sync 
licensing — negotiated terms declared in 
\texttt{agents.txt}, with fees proportional to 
the commercial value derived. Multi-user aggregation 
— where a single AI system serves millions of users 
using content from many publishers — requires a 
collective model analogous to music streaming 
royalties, discussed in 
Section~\ref{sec:economics}.

\subsection{The Open/Closed Source Analogy}
\label{sec:economics:opensource}

The intent-based tier model preserves a critical 
property of the current web: economic diversity. 
The web's strength has never derived from a single 
economic model — it derives from the coexistence 
of free and paid content, open and proprietary 
knowledge, advertising-supported and 
subscription-supported publishing. This diversity 
must be preserved in the agent-first web.

We propose framing the content economics decision 
for publishers through an analogy that has already 
proven robust in the adjacent domain of software: 
the open source / closed source distinction 
\cite{raymond1999cathedral}. Just as a software 
developer chooses whether to release code under 
an open license (freely available, potentially 
with attribution requirements) or a proprietary 
license (paid access, restricted use), a publisher 
chooses whether to make content available to agents 
under an open model or a paid model. The choice is 
entirely the publisher's — the framework mandates 
no universal model.

Under this framing:

\textbf{Open content} is freely accessible to 
agents under the same terms as humans. Rate limits 
apply as the boundary of free access — analogous 
to open source licenses that permit free use but 
not unlimited commercial exploitation. Wikipedia, 
government data, open research, personal blogs — 
all of these continue to operate as they do today, 
with agents welcome on the same terms as humans.

\textbf{Paid content} requires economic exchange 
for agent access. Publishers declare their pricing 
terms in \texttt{agents.txt} and content headers. 
The economic model — subscription, per-token, 
outcome-based — is chosen by the publisher. The 
agent checks declared terms before accessing and 
pays accordingly. The New York Times, academic 
journals, professional data services — all of 
these can operate paid agent access tiers 
independently of their human subscription models.

This framing is politically important as well as 
technically clean. It does not mandate that all 
content be paid — which would kill open knowledge 
and information access. It does not mandate that 
content be free — which would destroy publisher 
economics. It gives publishers the same choice 
software developers have had for decades and lets 
the market produce diversity.

\subsection{Token-Based Subscription Model}
\label{sec:economics:tokens}

For paid content, we propose the 
\textit{token-based subscription model} as the 
primary economic mechanism for agent content access. 
Rather than metering access by pageviews, article 
counts, or time periods — all human-centric units 
— content access is metered in tokens consumed, 
directly mirroring how AI API access is priced 
today \cite{anthropic2024mcp}.

Under this model, a publisher declares a price per 
million tokens for their content. An agent consuming 
that content is billed based on tokens actually 
consumed from the response. Publishers can declare 
different prices for different content tiers — 
breaking news vs archived content, premium analysis 
vs commodity information — using the 
\texttt{agents.txt} pricing declarations introduced 
in Section~\ref{sec:access:agentstxt}.

\begin{figure}[H]
\centering
\includegraphics[width=\textwidth]{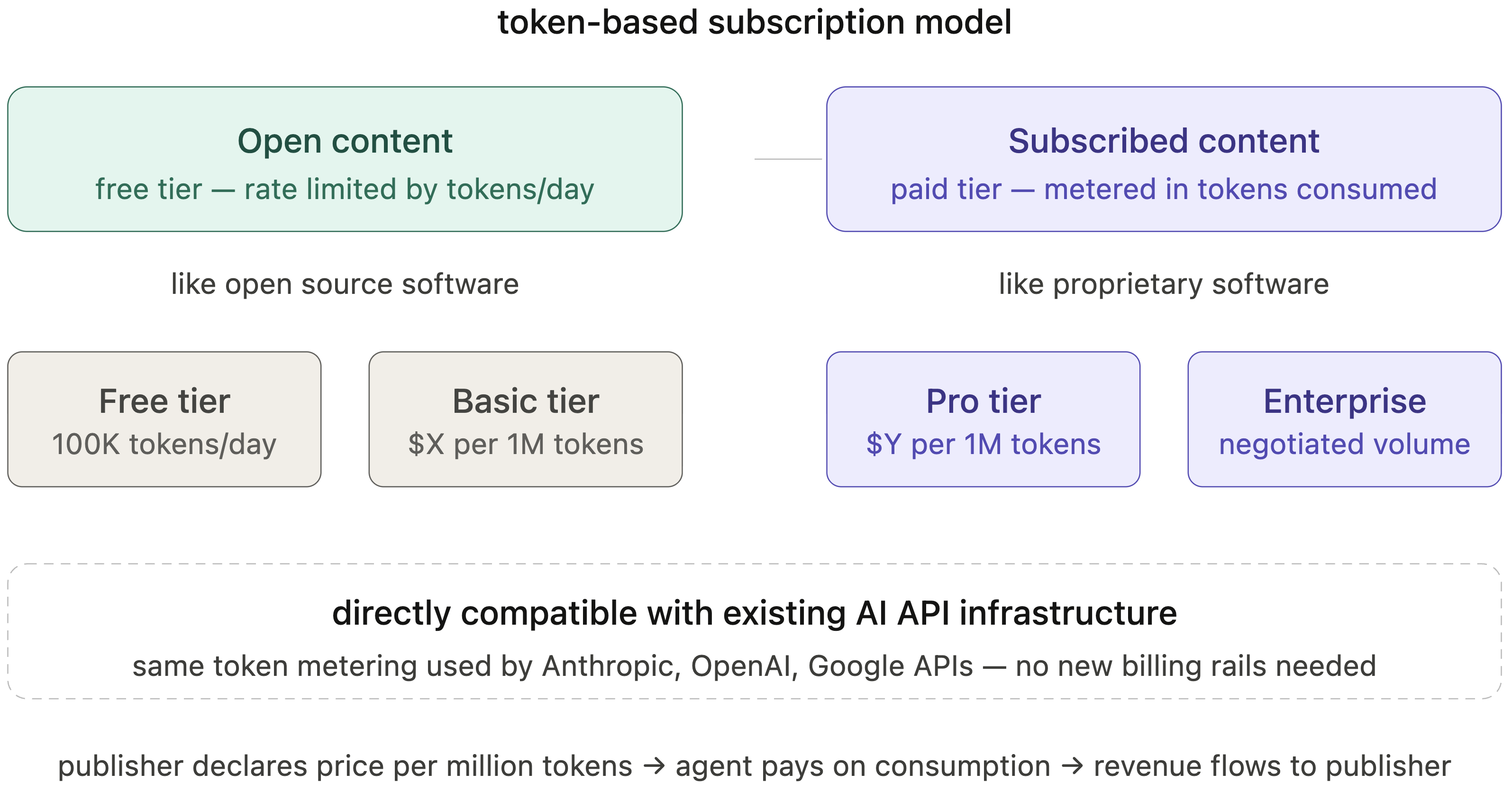}
\caption{Token-based subscription model. Content 
access is metered in tokens consumed rather than 
pageviews or article counts. Publishers declare 
price per million tokens; agents pay on consumption. 
Free tier access is rate-limited by token budget 
rather than blocked. The model is directly compatible 
with existing AI API billing infrastructure — the 
same token metering used by major AI providers 
requires no new payment rails for adoption.}
\label{fig:tokenmodel}
\end{figure}

Figure~\ref{fig:tokenmodel} illustrates the 
token-based subscription model and its relationship 
to existing AI infrastructure.

The token-based model has several properties that 
make it well-suited to agent content economics. 
First, it is directly compatible with existing AI 
billing infrastructure. AI providers already meter 
token consumption for API access — the same 
infrastructure can be extended to meter content 
consumption, requiring no new payment rails. Second, 
it is proportional to actual consumption. A short 
agent query that retrieves a single paragraph costs 
less than a deep research task that retrieves 
thousands of articles — unlike subscription models 
that charge fixed fees regardless of actual use. 
Third, it is format-agnostic. ATML content, 
Markdown content, structured data — all can be 
metered in tokens regardless of their format, 
providing a unified pricing mechanism across the 
heterogeneous content landscape of the web.

For free-tier access, the token budget functions 
as a rate limit rather than a paywall — an agent 
consuming 100,000 tokens per day of a publisher's 
content at no charge is welcome; an agent consuming 
10 million tokens per day for commercial purposes 
is directed to the paid tier. The boundary is 
declared in \texttt{agents.txt}, enforced by the 
agent identification metadata mechanism, and 
transparent to both agents and publishers.

\subsection{The Commissioned Content Economy}
\label{sec:economics:commissioned}

The economic models described above address the 
demand side of agent content economics — how agents 
pay for content they consume. We now address the 
supply side: how content production in the 
agent-first web can be economically sustainable 
and epistemically grounded.

The epistemic recursion problem identified in 
Section~\ref{sec:diagnosis:content} — the 
self-referential loop in which AI generates content 
that AI consumes — arises precisely because 
AI-generated content currently carries no economic 
cost and no provenance requirement. Content farms 
flood the web with AI-generated articles at 
negligible marginal cost; there is no economic 
signal distinguishing this content from 
human-authored journalism, and no provenance 
mechanism flagging it as AI-derived. The loop is 
sustained by the absence of economic friction at 
the production stage.

We propose the \textit{commissioned content economy} 
as a structural solution to this problem. Under 
this model, AI-generated content that enters the 
web as a publishable resource — rather than as an 
ephemeral agent response — is produced under a 
commissioning relationship: a human entity pays 
for the content to be produced, establishing a 
human intentionality anchor at the production stage. 
The commissioned content is published with a 
machine-readable provenance tag declaring the 
commissioner, the producing agent, and the human 
oversight level. Other agents and humans can access 
the content under declared terms, with revenue 
flowing back to the commissioner.

\begin{figure}[H]
\centering
\includegraphics[width=\textwidth]{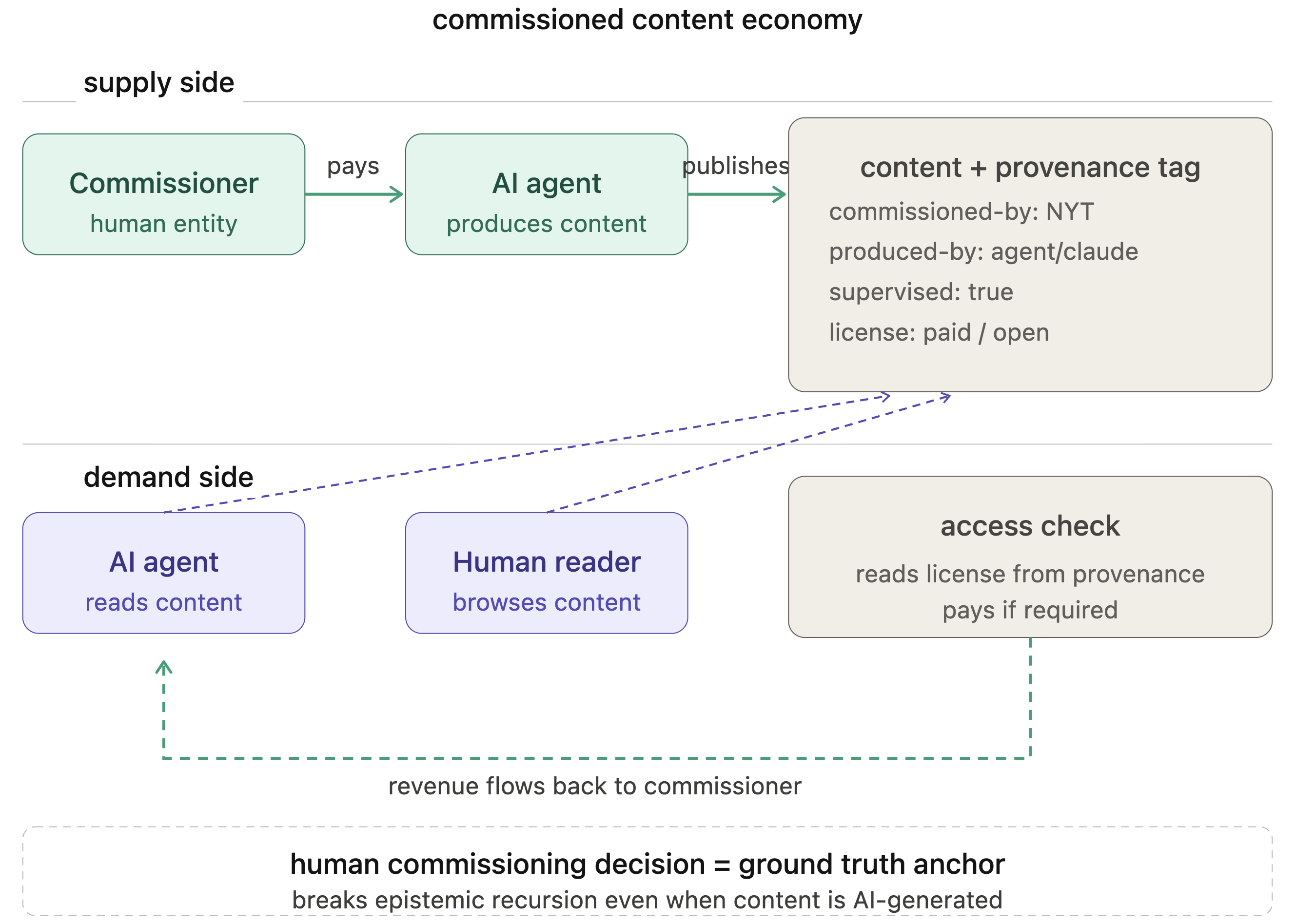}
\caption{The commissioned content economy. On the 
supply side, a human entity commissions an AI agent 
to produce content under a paid agreement. The 
content is published with a provenance tag declaring 
commissioner, producing agent, and human supervision 
level. On the demand side, other agents and human 
readers access the content under declared terms, 
paying where required. Revenue flows back to the 
commissioner. The critical contribution of this 
model is the human commissioning decision at the 
production stage — which introduces human 
intentionality as a ground truth anchor, breaking 
the epistemic recursion loop even when the content 
itself is AI-generated.}
\label{fig:commissioned}
\end{figure}

Figure~\ref{fig:commissioned} illustrates the 
commissioned content economy as a two-sided 
marketplace.

The commissioned content model does not require that 
all web content be commissioned — the open web 
continues to exist for informal, personal, and 
community-generated content. It provides an economic 
structure for the subset of web content where quality, 
provenance, and trustworthiness matter most: 
journalism, research, professional analysis, 
educational content. For this subset, the 
commissioning relationship reintroduces the human 
judgment that the epistemic recursion loop erodes 
— not by requiring humans to write every word, 
but by requiring a human to decide that this 
content is worth producing and to take 
responsibility for it.

\subsection{The Agent Advertising Model}
\label{sec:economics:advertising}

The attention economy's primary mechanism — 
advertising — does not disappear in the agent-first 
web. It transforms. When an agent retrieves content 
on behalf of a human user and presents a synthesized 
response, there remains a human attention moment: 
the user reading the agent's output. This moment 
is potentially monetizable, though the mechanism 
differs fundamentally from traditional web 
advertising.

We sketch a preliminary model for agent-targeted 
advertising, acknowledging that full development 
of this model constitutes a separate research 
contribution beyond the scope of this paper. Under 
the proposed model, publishers embed structured 
advertisement metadata in their ATML content 
alongside substantive content. When an agent 
retrieves and synthesizes this content, it carries 
the advertisement metadata as part of its response 
payload. The agent frontend — Claude, ChatGPT, 
Gemini, or other interface — surfaces the 
advertisement to the human user at an appropriate 
point in the interaction. An impression is recorded 
when the advertisement reaches the human; revenue 
flows to the publisher whose content carried it.

This model faces a significant open challenge: agent 
frontends have limited incentive to surface publisher 
advertisements, particularly when doing so competes 
with their own monetization interests. Resolving 
this tension requires either a protocol-level 
standard mandating advertisement carriage — 
analogous to the must-carry rules in broadcast 
television regulation — or a revenue-sharing 
agreement between publishers and agent frontend 
operators. Both approaches involve stakeholder 
coordination that extends beyond technical 
standardization. We identify this as a priority 
open challenge in Section~\ref{sec:challenges} 
and flag it as a direction for future work.

\subsection{Summary}
\label{sec:economics:summary}

Table~\ref{tab:economics} summarizes the economic 
layer redesign proposals and their relationship to 
the failures diagnosed in 
Section~\ref{sec:diagnosis:economics}.

\begin{table}[H]
\centering
\caption{Economic layer redesign: diagnosed failures 
and proposed solutions.}
\label{tab:economics}
\begin{adjustbox}{width=0.48\textwidth}
\begin{tabular}{p{3cm}p{4.5cm}}
\toprule
\thead{Diagnosed failure} & 
\thead{Proposed solution} \\
\midrule
Attention economy collapses under agent interaction &
Intent-based tier model grounded in agent-as-human-proxy principle \\
\addlinespace
No value exchange at point of agent consumption &
Token-based subscription model — metered on consumption \\
\addlinespace
Universal payment mandates kill open knowledge &
Open/closed source analogy — publisher chooses model \\
\addlinespace
No economic model for AI content production &
Commissioned content economy with human intentionality anchor \\
\addlinespace
Attribution impossible at scale &
Collective pool model for multi-user aggregation tiers \\
\addlinespace
Ad model breaks without human attention &
Agent advertising model (open challenge — future work) \\
\bottomrule
\end{tabular}
\end{adjustbox}
\end{table}

The economic layer redesign is the most complex 
of the three layers because it must accommodate 
the full diversity of the existing web — free and 
paid, open and proprietary, advertising-supported 
and subscription-supported — while introducing 
new mechanisms for the genuinely novel cases that 
agent interaction creates. The intent-based tier 
model, grounded in the agent-as-human-proxy 
principle, provides the organizing framework. 
The token-based subscription model provides the 
primary mechanism for paid access. The commissioned 
content economy addresses the supply side. Together 
they constitute a coherent replacement for the 
attention economy — one designed from first 
principles for a web in which agents are primary 
consumers of content.

\section{The Content Layer Redesign}
\label{sec:content}

The access layer redesign (Section~\ref{sec:access}) 
establishes how agents reach content and identify 
themselves. The economic layer redesign 
(Section~\ref{sec:economics}) establishes how value 
flows when they do. This section addresses what the 
content itself looks like in an agent-first web — 
proposing a new content format, a human supervision 
standard, a provenance chain architecture, and an 
agent-native discoverability mechanism. Together 
these proposals address the three content-layer 
incompatibilities diagnosed in 
Section~\ref{sec:diagnosis:content}: format 
inefficiency, provenance absence, and epistemic 
recursion.

\subsection{Design Principles for Agent-First Content}
\label{sec:content:principles}

Before proposing specific mechanisms, we establish 
the design principles that agent-first content must 
satisfy. These principles are derived from the 
requirements of LLM-based agents as content 
consumers — requirements that are fundamentally 
different from those of human readers.

\textbf{Semantic richness over visual presentation.} 
Agents process meaning, not appearance. HTML's 
investment in visual layout — typography, color, 
responsive grid systems, animation — is not merely 
neutral overhead for agents; it is active noise that 
consumes token budget without contributing to 
comprehension. Agent-first content must prioritize 
semantic structure: what is being said, who said it, 
what it is derived from, and how confident the 
source is. Visual presentation is a secondary 
concern addressed by the rendering layer of the 
agent frontend, not the content layer.

\textbf{Explicit provenance as a first-class 
property.} An agent retrieving content must be 
able to determine its origin, derivation chain, 
and human oversight level without inferring it 
from context. Provenance is not metadata appended 
to content — it is a structural property of the 
content itself, as fundamental as the content body.

\textbf{Machine-readable access and license terms.} 
An agent must be able to determine, before 
consuming content, what terms govern that 
consumption — free or paid, personal or commercial, 
redistributable or restricted. These terms must be 
embedded in the content itself or in a well-known 
declaration file, not buried in a human-readable 
terms-of-service page that no agent can reliably 
parse.

\textbf{Token efficiency.} Every token an agent 
consumes processing layout noise rather than 
semantic content is a token wasted. 
\citet{webmcp2025, slice-mcp} demonstrated that current HTML 
carries 67.6\% token overhead relative to semantic 
content alone. Agent-first content must minimize 
this overhead — not merely for economic reasons 
but because token efficiency directly determines 
the quality and depth of agent reasoning over 
content at scale.

\textbf{Dual-layer compatibility.} The human web 
does not disappear. Agent-first content must 
coexist with human-readable content through the 
dual-layer architecture proposed in 
Section~\ref{sec:access:duallayer} — served from 
the same domain, produced from the same content 
repository, differentiated only at the delivery 
layer.

\subsection{ATML: Agent Text Markup Language}
\label{sec:content:atml}

We propose the \textit{Agent Text Markup Language 
(ATML)} as the content format for agent-first web 
delivery. ATML is not an entirely new language — 
it is a structured semantic profile designed for 
machine consumption, composed of three explicit 
layers: provenance, access terms, and semantic 
content. We engage first with the question of 
whether a new format is necessary at all.

\subsubsection{The HTML Debate}

\citet{karpathy2025html} has argued that HTML is 
already sufficiently structured for agent 
consumption — that modern LLMs can parse HTML 
effectively and that a new format is unnecessary 
overhead. This position has merit: HTML does 
encode semantic structure through heading tags, 
paragraph tags, list tags, and semantic HTML5 
elements such as \texttt{<article>}, 
\texttt{<section>}, and \texttt{<main>}. Many 
agents do navigate HTML-rendered web content 
successfully today.

We argue, however, that HTML is inadequate for 
three reasons beyond parsing difficulty. First, 
the token overhead problem is structural, not 
incidental: even well-structured semantic HTML 
carries layout, styling, and scripting overhead 
that is irrelevant to agent consumption. The 
67.6\% token reduction demonstrated by 
\citet{webmcp2025, nurolense} was achieved not by better 
HTML parsing but by delivering a semantically 
equivalent content representation without layout 
noise — a task HTML cannot accomplish without 
effectively becoming a different format. Second, 
HTML has no native provenance layer. The 
information an agent most needs — who authored 
this, what was it derived from, what supervision 
level applies — has no standardized HTML 
representation. Third, HTML has no native 
machine-readable access terms layer. An agent 
cannot read a content license from HTML 
structure; it must parse a separate legal 
document written for human readers.

ATML addresses all three gaps while remaining 
grounded in familiar XML syntax that existing 
tooling can process without new parsers.

\subsubsection{ATML Structure}

\begin{figure}[H]
\centering
\includegraphics[width=\textwidth]{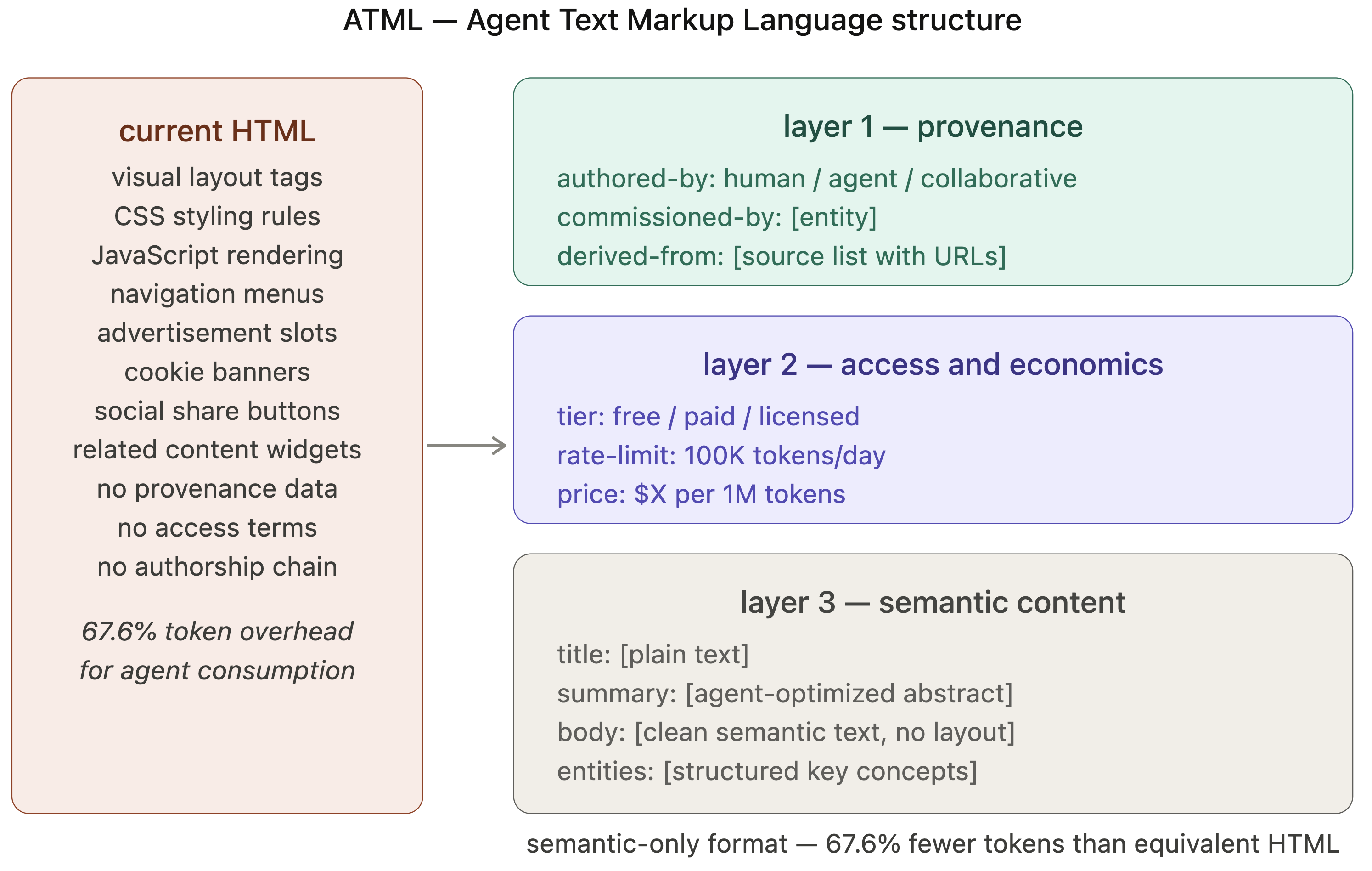}
\caption{ATML structure compared to current HTML. 
HTML carries significant visual rendering overhead 
— layout tags, CSS, JavaScript, navigation, 
advertisement slots — that is irrelevant to agent 
consumption and imposes a 67.6\% token overhead 
\cite{webmcp2025}. ATML replaces this with three 
explicit semantic layers: provenance (declaring 
origin, commissioner, derivation chain, and human 
supervision level), access and economics (declaring 
tier, rate limits, and pricing), and semantic 
content (clean text, summary, and structured 
entities). The result is a format optimized for 
agent comprehension while carrying all the 
metadata agents need to evaluate, cite, and pay 
for content appropriately.}
\label{fig:atml}
\end{figure}

ATML is organized into three layers, as 
illustrated in Figure~\ref{fig:atml}.

A minimal ATML document has the following 
structure:

\begin{verbatim}
<atml version="1.0">
  <provenance>
    <authored-by>human</authored-by>
    <commissioned-by>NYT</commissioned-by>
    <derived-from>
      <source url="https://..." 
              retrieved="2025-11-01"/>
    </derived-from>
    <supervised>true</supervised>
    <supervision-level>2</supervision-level>
  </provenance>
  <access>
    <tier>paid</tier>
    <rate-limit unit="tokens" 
                period="day">100000</rate-limit>
    <price unit="per-million-tokens">
      2.50
    </price>
    <license>personal-use</license>
    <license>no-training</license>
  </access>
  <content>
    <title>Article title here</title>
    <summary>Agent-optimized abstract 
             in 2-3 sentences.</summary>
    <body>Clean semantic text without 
          layout or styling markup.
    </body>
    <entities>
      <entity type="person">Name</entity>
      <entity type="org">Organisation
      </entity>
    </entities>
  </content>
</atml>
\end{verbatim}

The \texttt{<provenance>} layer is the ATML 
element with no HTML equivalent. It declares the 
authorship type (human, agent, or collaborative), 
the commissioning entity, the sources the content 
was derived from with retrieval dates, and the 
human supervision level from the tier model 
proposed in Section~\ref{sec:content:supervision}. 
This layer directly enables agents to make 
quality-aware content decisions and breaks the 
epistemic recursion loop by making AI-origin 
content explicitly visible.

The \texttt{<access>} layer declares economic 
terms in machine-readable form, directly 
compatible with the token-based subscription 
model proposed in 
Section~\ref{sec:economics:tokens}. Agents read 
these terms before consuming content — no 
separate legal document required.

The \texttt{<content>} layer carries the semantic 
text without layout noise. The \texttt{<summary>} 
field enables agents to make relevance decisions 
before consuming the full body — a significant 
token efficiency gain for research tasks 
involving many sources. The \texttt{<entities>} 
field provides structured key concepts that 
enable downstream reasoning without full-body 
parsing.

\subsection{Human Supervision Requirements}
\label{sec:content:supervision}

The epistemic recursion problem identified in 
Section~\ref{sec:diagnosis:content} cannot be 
solved by format alone. A provenance tag that 
always declares \texttt{<supervised>false</supervised>} 
for AI-generated content is honest but does not 
break the recursion loop — it merely labels it. 
Breaking the loop requires reintroducing human 
intentionality into the content production chain 
at a structural level.

\begin{figure}[H]
\centering
\includegraphics[width=\textwidth]{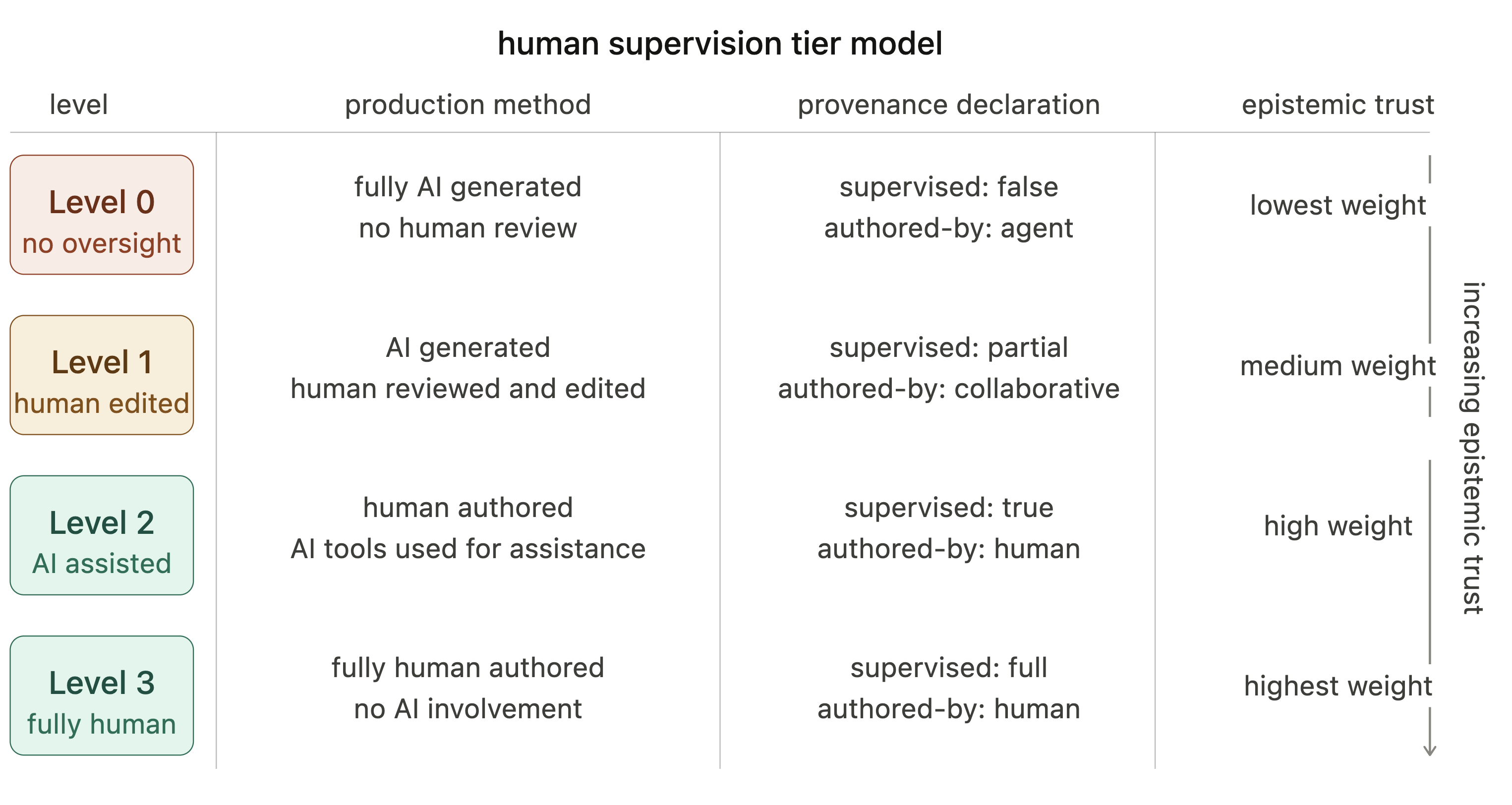}
\caption{Human supervision tier model. Content is 
classified into four levels based on degree of 
human involvement in production, from Level 0 
(fully AI generated, no human review) to Level 3 
(fully human authored). Each level maps to a 
declared ATML provenance value and carries 
implications for epistemic trust — the weight 
agents and search engines assign to the content 
in reasoning and ranking. The model does not 
prohibit AI-generated content; it makes the degree 
of human oversight transparent, enabling agents 
to make quality-aware consumption decisions.}
\label{fig:supervision}
\end{figure}

We propose a four-level \textit{human supervision 
tier model} that classifies content by the degree 
of human involvement in its production. Each level 
maps to a declared provenance value in ATML and 
carries implications for how agents should weight 
the content in their reasoning. Figure~\ref{fig:supervision} 
illustrates the model.

\textbf{Level 0 — No human oversight.} Content 
is fully AI generated with no human review before 
publication. Declared as \texttt{supervised: false} 
in ATML provenance. Agents should assign lowest 
epistemic weight to Level 0 content and avoid 
using it as a primary source for factual claims. 
Search engines and agent content indices should 
rank Level 0 content below human-supervised 
content for informational queries.

\textbf{Level 1 — Human edited.} Content is 
AI generated but reviewed and edited by a human 
before publication. The human editor takes 
responsibility for factual accuracy and quality. 
Declared as \texttt{supervised: partial}. 
Represents the minimum acceptable supervision 
level for content that will be cited by agents 
as a source.

\textbf{Level 2 — AI assisted.} Content is 
human authored with AI tools used for assistance 
— drafting, research, editing — but with a human 
as the primary author and decision-maker. 
Declared as \texttt{supervised: true, authored-by: 
human}. This level encompasses most professional 
content production in the near-term future and 
should be treated as equivalent to traditional 
human-authored content.

\textbf{Level 3 — Fully human authored.} Content 
is produced without AI involvement. Declared as 
\texttt{supervised: full, authored-by: human}. 
This level provides the strongest epistemic 
ground truth signal and should receive highest 
weight in agent reasoning for factual claims.

The supervision tier model does not prohibit 
any level of AI involvement in content 
production. It makes the degree of human 
oversight transparent and machine-readable, 
enabling agents to make quality-aware content 
decisions rather than treating all web content 
as equally trustworthy regardless of origin. 
This is the structural mechanism by which the 
web can sustain a diverse content ecosystem — 
including AI-generated content — without 
collapsing into the epistemic recursion loop.

\subsection{Provenance Chain Architecture}
\label{sec:content:provenance}

The supervision tier model establishes what level 
of human oversight applies to a piece of content. 
The \textit{provenance chain architecture} 
establishes how that claim is verified. A 
provenance declaration is only as useful as the 
trust placed in it — an AI content farm could 
declare \texttt{supervised: true} and break the 
epistemic intent of the model entirely. 
Cryptographic verification is required.

We propose building ATML provenance on the 
Coalition for Content Provenance and Authenticity 
(C2PA) standard \cite{c2pa2024}, extending it 
with agent-specific fields. Under this 
architecture, each ATML document carries a 
cryptographic provenance certificate that 
encodes:

\begin{itemize}
    \item The identity of the publishing entity, 
    verified against a public key infrastructure
    \item The supervision level declared, signed 
    by the publishing entity
    \item A hash of the content body at time of 
    publication, enabling verification that 
    content has not been modified post-publication
    \item A chain of derived-from references, 
    each with its own provenance certificate, 
    enabling agents to trace the epistemic 
    lineage of a claim back to primary sources
\end{itemize}

This architecture does not guarantee that 
supervision declarations are honest — a bad 
actor who controls their own signing key can 
still declare false supervision levels. However, 
it makes false declarations attributable: if a 
publisher is found to have systematically 
declared false supervision levels, their signing 
key can be revoked from the public key 
infrastructure, removing trust from all their 
content. The reputational and legal consequences 
of provenance fraud create a deterrent analogous 
to the consequences of false advertising in 
existing legal frameworks.

The derived-from chain is particularly important 
for breaking epistemic recursion. When an agent 
traces the provenance chain of a piece of 
content and finds that it derives from another 
AI-generated piece, which derives from another 
AI-generated piece, without a human-authored 
primary source anywhere in the chain, it has 
detected epistemic recursion in action. Agents 
can be designed to flag such chains and reduce 
the weight assigned to content with deep 
AI-only derivation chains — a practical circuit 
breaker for the recursion loop that requires 
no centralised authority to operate.

\subsection{Agent Content Discoverability}
\label{sec:content:discovery}

The human web's discovery mechanism — search 
engines optimizing for human attention and 
click-through — does not serve agent needs. An 
agent does not want the most popular result; 
it wants the most authoritative, most 
accurately provenance-tagged, most 
appropriately-priced result for its specific 
task. These are different optimization targets 
requiring different infrastructure.

\begin{figure}[H]
\centering
\includegraphics[width=\textwidth]{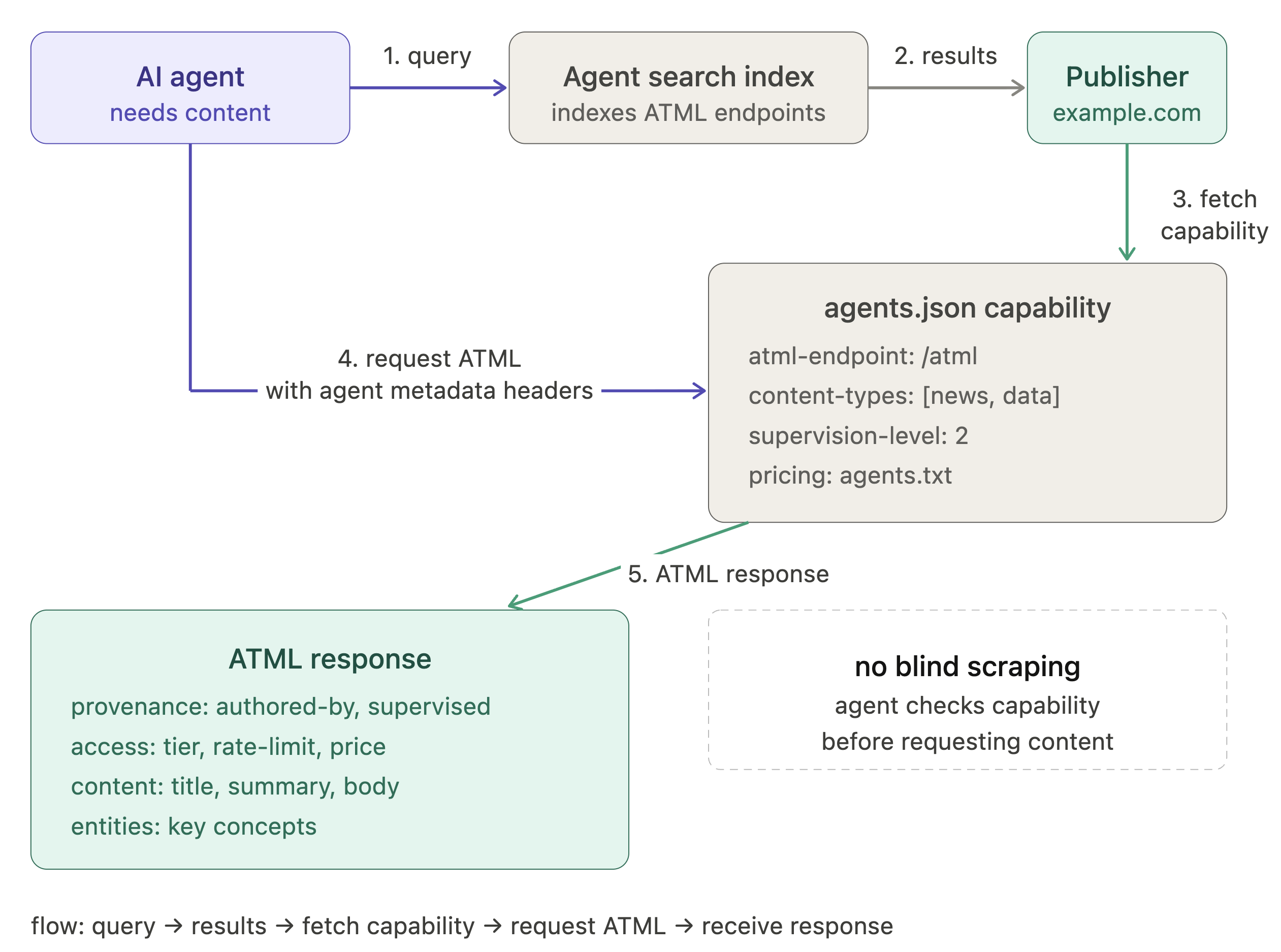}
\caption{Agent content discoverability 
architecture. An agent queries an agent search 
index optimized for semantic relevance, 
provenance quality, and access terms rather than 
human attention signals. The index returns ATML 
endpoints from publishers who have declared 
agent capabilities in a structured 
\texttt{agents.json} file. The agent fetches the 
capability declaration to understand available 
content types, supervision levels, and pricing 
before making content requests. This eliminates 
blind scraping — agents always know what they 
are accessing and under what terms before they 
access it.}
\label{fig:discovery}
\end{figure}

We propose an \textit{agent content 
discoverability architecture} built on a 
capability declaration standard and an 
agent-optimized search index. Figure~\ref{fig:discovery} 
illustrates the full discoverability flow.

The capability declaration — served at 
\texttt{/.well-known/agents.json} — extends 
the \texttt{agents.txt} access policy file 
proposed in Section~\ref{sec:access:agentstxt} 
with richer content metadata:

\begin{verbatim}
{
  "atml-endpoint": "/atml",
  "content-types": ["news", "analysis", 
                    "data"],
  "languages": ["en", "fr"],
  "supervision-level": 2,
  "auth": "delegation-token",
  "pricing": "/.well-known/agents.txt",
  "search-topics": ["finance", 
                    "technology"],
  "update-frequency": "real-time"
}
\end{verbatim}

Agent search engines index these capability 
declarations rather than raw content — they 
know which sites publish which content types 
at which supervision levels and under what 
economic terms, without having to scrape and 
parse the full content of every page. This 
is a fundamentally more efficient discovery 
architecture than human-web search, which 
must index full page content to infer topical 
relevance.

The agent search index optimizes for different 
signals than human search: semantic relevance 
to agent task, provenance quality of the 
source, supervision level of available content, 
and access terms compatibility with the agent's 
authorization level. These signals produce 
rankings appropriate for agent information 
retrieval tasks — prioritizing trustworthy, 
well-provenance-tagged sources over 
high-traffic, SEO-optimized pages.

\subsection{Dual Publishing Strategy}
\label{sec:content:dualpublish}

The transition from a human-centric to an 
agent-first content layer cannot be abrupt. 
Publishers cannot abandon HTML overnight; 
human users continue to browse the web 
directly; legacy systems depend on existing 
content formats. The dual-layer web 
architecture proposed in 
Section~\ref{sec:access:duallayer} requires 
a corresponding dual publishing strategy at 
the content layer.

We propose that publishers adopt ATML as a 
parallel delivery format alongside HTML, 
generated from the same content source. In 
most publishing systems, content is stored 
in a structured internal format — a content 
management system database — and rendered 
into HTML at delivery time. ATML is an 
alternative rendering target for the same 
source content, not a separate authoring 
workflow. A publisher who stores articles 
as structured database records can generate 
both an HTML rendering (for browsers) and 
an ATML rendering (for agents) from the 
same record, adding only the provenance 
and access layers that ATML requires and 
HTML does not.

The migration path mirrors the access layer 
migration described in 
Section~\ref{sec:access:duallayer}: Phase 1 
establishes ATML as an optional parallel 
format; Phase 2 normalizes dual publishing 
as standard practice; Phase 3 establishes 
ATML as the primary format for agent 
consumption with HTML maintained for 
browser compatibility. The human web does 
not disappear — it becomes a rendering 
target for a content layer that is 
fundamentally structured for agent 
consumption.

\subsection{Summary}
\label{sec:content:summary}

Table~\ref{tab:content} summarizes the content 
layer redesign proposals and their relationship 
to the failures diagnosed in 
Section~\ref{sec:diagnosis:content}.

\begin{table}[H]
\centering
\caption{Content layer redesign: diagnosed 
failures and proposed solutions.}
\label{tab:content}
\begin{adjustbox}{width=0.48\textwidth}
\begin{tabular}{p{3cm}p{4.5cm}}
\toprule
\thead{Diagnosed failure} & 
\thead{Proposed solution} \\
\midrule
HTML format overhead — 67.6\% token waste &
ATML — semantic content format without layout noise \\
\addlinespace
No provenance standard &
ATML provenance layer + C2PA cryptographic chain \\
\addlinespace
No human supervision signal &
Four-level human supervision tier model \\
\addlinespace
Epistemic recursion — AI feeding AI &
Provenance chain circuit breaker + supervision declaration \\
\addlinespace
No agent discoverability standard &
agents.json capability declaration + agent search index \\
\addlinespace
Human-only publishing workflow &
Dual publishing strategy — HTML and ATML from same source \\
\bottomrule
\end{tabular}
\end{adjustbox}
\end{table}

The content layer redesign closes the loop 
opened by the access and economic layer 
redesigns. An agent that can identify itself 
(Section~\ref{sec:access}), operate under 
an appropriate economic model 
(Section~\ref{sec:economics}), and retrieve 
semantically rich, provenance-tagged content 
in an efficient format (this section) has 
everything it needs to operate as a 
first-class web citizen. The three layers 
are deeply interdependent — ATML content 
is only useful if agents can reach it 
(access layer) and publishers have an 
incentive to produce it (economic layer) 
— which is why all three must be redesigned 
simultaneously rather than in isolation.

\section{The Agent-First Web: A Unified Framework}
\label{sec:framework}

The preceding three sections proposed solutions to 
each layer of the human-centric web's failure under 
agent interaction: an access layer redesign 
(Section~\ref{sec:access}), an economic layer 
redesign (Section~\ref{sec:economics}), and a 
content layer redesign (Section~\ref{sec:content}). 
This section synthesizes these proposals into a 
unified framework — establishing the design 
principles, social contract restatement, and 
migration roadmap that constitute the paper's 
headline contribution.

\subsection{The Social Contract Restatement}
\label{sec:framework:contract}

The web's original social contract — articulated 
implicitly through the architectural choices of 
its first decade — rested on three foundational 
commitments \cite{berners1989information, 
zittrain2008future}. First, open access: any 
client should be able to retrieve any resource 
without prior negotiation or identity disclosure. 
Second, the attention economy: publishers provide 
content freely in exchange for human attention 
monetized through advertising. Third, human 
authorship: the web is a record of human 
knowledge, experience, and expression, authored 
by humans for human readers.

Each of these commitments is now under structural 
pressure. Open access is being dismantled by 
blanket agent blocking. The attention economy is 
collapsing under zero-click AI synthesis. Human 
authorship is being displaced by AI-generated 
content at scale. The web's social contract is 
not merely being stressed — it is being violated 
simultaneously at all three layers by forces its 
architects did not anticipate.

\begin{figure}[H]
\centering
\includegraphics[width=\textwidth]{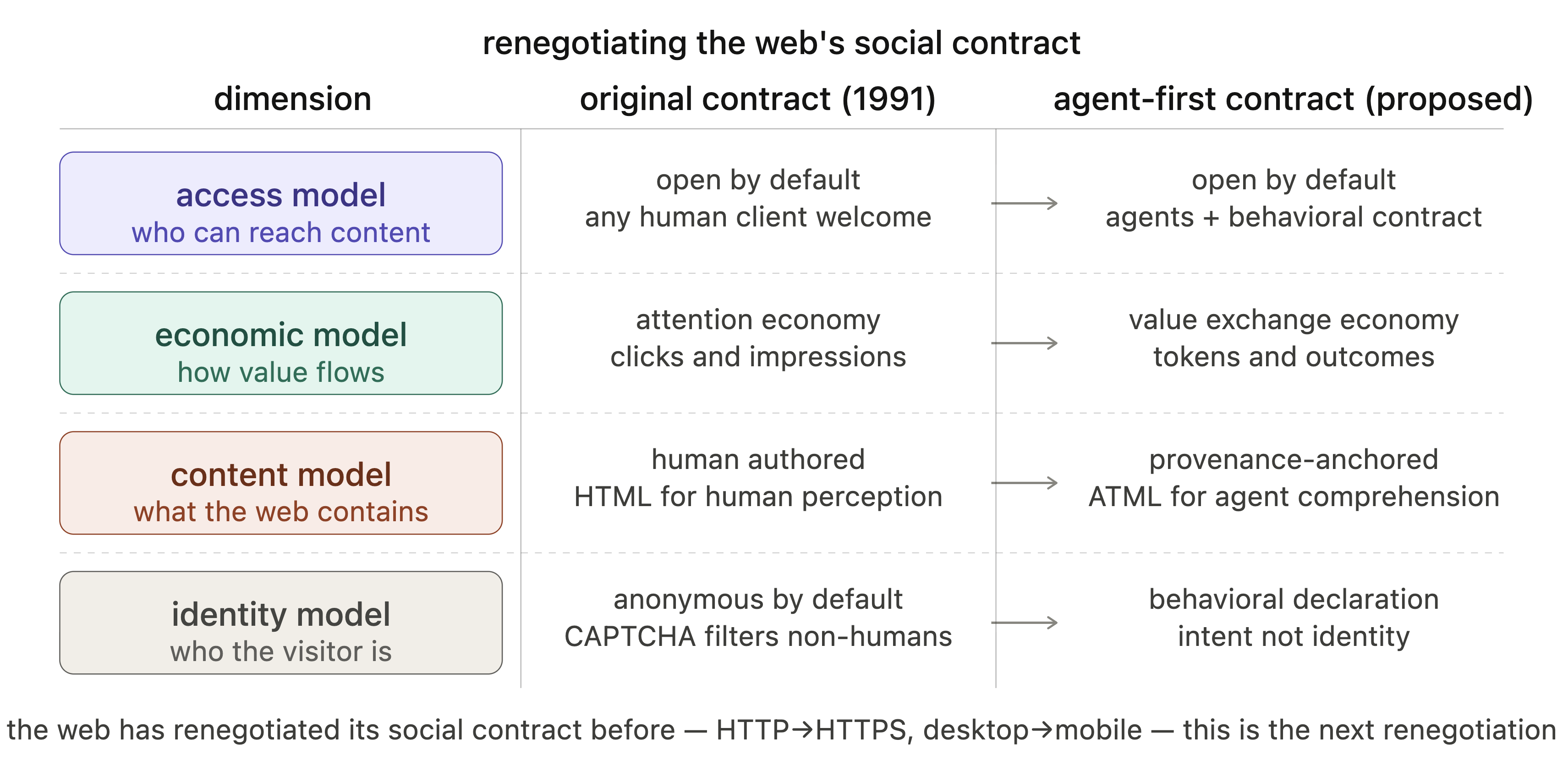}
\caption{The web's social contract: original 
versus agent-first. The original contract 
(1991) rested on open access for human 
clients, an attention economy, HTML for 
human perception, and anonymous identity. 
The agent-first contract preserves open 
access but extends it to agents through 
behavioral contracts; replaces the attention 
economy with a value-exchange economy based 
on tokens and outcomes; replaces HTML with 
provenance-anchored ATML; and replaces 
CAPTCHA-based human filtering with 
intent-based behavioral declaration. The 
web has renegotiated its social contract 
before — HTTP to HTTPS, desktop to mobile 
— this is the next renegotiation.}
\label{fig:contract}
\end{figure}

We propose a restatement of the web's social 
contract for the agent-first era, preserving 
its founding values while extending them to 
accommodate the new realities of agent-mediated 
interaction. Figure~\ref{fig:contract} 
illustrates the comparison between the original 
and proposed contracts.

The new social contract can be stated 
concisely:

\begin{quote}
\textit{The agent-first web is open to any 
client — human or agent — that declares its 
behavioral context and respects the terms 
declared by content providers. Value flows 
at the point of consumption, not the point 
of attention. Content carries its own 
provenance and human oversight level, 
enabling quality-aware consumption. The 
human web continues to exist; the agent 
web is built alongside it, not in 
replacement of it.}
\end{quote}

This restatement preserves the web's founding 
commitment to openness — no client is excluded 
by identity — while introducing the behavioral 
accountability and economic transparency that 
agent-scale interaction requires.

\subsection{The Ten Principles of an 
             Agent-First Internet}
\label{sec:framework:principles}

We synthesize the proposals of 
Sections~\ref{sec:access}--\ref{sec:content} 
into ten design principles for the agent-first 
internet. These principles are intended to 
serve the same function that REST principles 
served for API design \cite{fielding2000rest} 
or mobile-first principles served for 
responsive web design \cite{marcotte2011responsive}: 
a concise, quotable framework that practitioners 
can apply when making architectural decisions 
about agent-web interaction.

\begin{enumerate}

\item \textbf{Presumption of agent access.} 
Agents acting on behalf of humans are presumed 
welcome on any web resource that is open to 
anonymous human visitors. Blocking is reserved 
for malicious behavior, not agent identity. 
The default posture is access, not exclusion.

\item \textbf{Behavioral contract over identity 
mandate.} Agents declare their behavioral 
context — who they represent, at what scale, 
for what purpose — rather than proving their 
identity. Anonymous is acceptable; unaccountable 
is not. Intent declaration enables graduated 
responses without requiring privacy-invasive 
identification.

\item \textbf{Rate limiting as the universal 
access control.} The binary block/allow 
decision is replaced by graduated rate 
limiting determined by declared intent and 
authentication level. No legitimate agent 
use case is categorically blocked; all are 
accommodated at appropriate scale and under 
appropriate terms.

\item \textbf{Economic equivalence to the 
represented human.} An agent's economic 
obligation mirrors that of the human it 
represents. Free content for anonymous humans 
is free for their agents. Paid subscriptions 
extend to agents through delegation tokens. 
New economic models are required only for 
genuinely novel agent behaviors with no 
human analog.

\item \textbf{Token as the unit of value 
exchange.} The pageview and the advertising 
impression are replaced by the token as the 
fundamental unit of value exchange between 
agents and content providers. Token 
consumption is proportional to actual value 
derived, compatible with existing AI billing 
infrastructure, and format-agnostic.

\item \textbf{Publisher economic sovereignty.} 
No universal economic model is mandated. 
Publishers choose between open and paid 
content — mirroring the open source / 
proprietary software distinction — and 
declare their terms in machine-readable 
form. Economic diversity is preserved as 
a web property.

\item \textbf{Provenance as a first-class 
web property.} Every piece of web content 
carries a machine-readable provenance 
declaration: who authored it, what it was 
derived from, and what degree of human 
oversight was applied. Provenance is not 
optional metadata — it is a structural 
property of agent-first content, as 
fundamental as the content body itself.

\item \textbf{Human supervision declared, 
not assumed.} The degree of human 
involvement in content production is 
declared explicitly through the supervision 
tier model rather than inferred or assumed. 
Agents weight content in their reasoning 
according to declared supervision level. 
AI-generated content is not prohibited; 
it is labelled.

\item \textbf{Dual-layer coexistence.} The 
human web and the agent web coexist on the 
same infrastructure. HTML for human 
browsers and ATML for agent clients are 
served from the same domain, generated 
from the same content source, differentiated 
only at the delivery layer. Migration is 
gradual and additive, not disruptive.

\item \textbf{Simultaneous three-layer 
redesign.} Access, economics, and content 
are redesigned together. Fixing any single 
layer in isolation produces solutions that 
are undermined by the failures of the other 
two. The three layers are deeply coupled 
— they must be addressed as a system, not 
as independent problems.

\end{enumerate}

\begin{figure}[H]
\centering
\includegraphics[width=\textwidth]{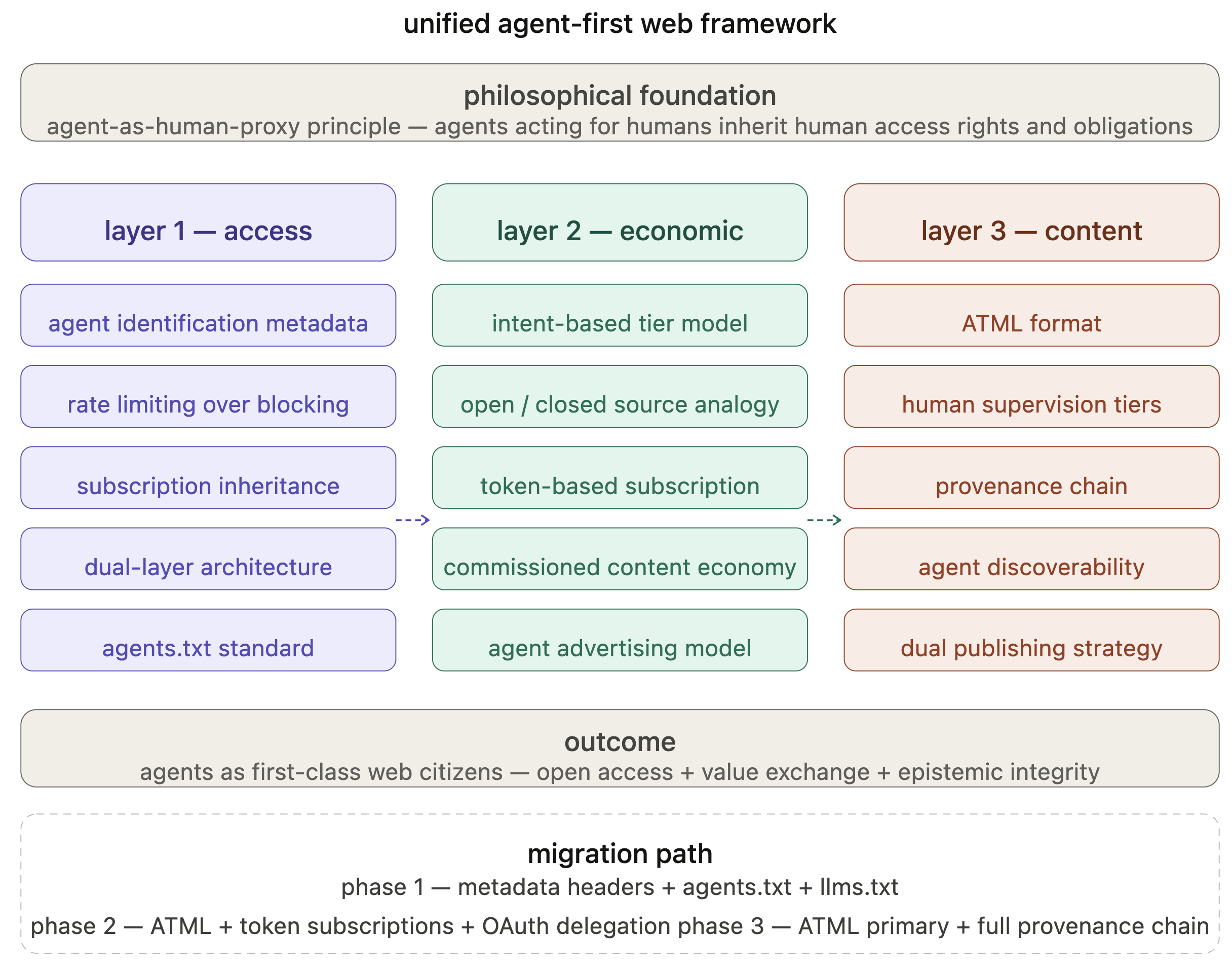}
\caption{The unified agent-first web 
framework. The agent-as-human-proxy 
principle provides the philosophical 
foundation. Three redesign layers — access, 
economic, and content — each contribute 
specific mechanisms. Together they produce 
an agent-first internet in which agents 
are first-class citizens with open access, 
value-exchange economics, and 
provenance-anchored content. The migration 
path is phased and additive, preserving 
the human web throughout the transition.}
\label{fig:framework}
\end{figure}

These ten principles constitute the 
\textit{agent-first internet framework}. 
Figure~\ref{fig:framework} illustrates how 
the principles map to the three layers and 
their interdependencies.

\subsection{Migration Roadmap}
\label{sec:framework:migration}

The agent-first web is not built overnight. 
The proposals in this paper span a range of 
implementation complexity — from the 
immediately deployable (agent metadata 
headers, \texttt{agents.txt}) to the 
medium-term (ATML adoption, OAuth delegation 
tokens) to the longer-term (full provenance 
chain infrastructure, agent search indices). 
We propose a three-phase migration roadmap 
that sequences these proposals by 
implementation complexity and stakeholder 
readiness.

\textbf{Phase 1 — Identification and 
Declaration (near-term).} The immediate 
priority is establishing the identification 
and declaration infrastructure that enables 
all subsequent proposals. This includes: 
deployment of agent metadata HTTP headers 
by major AI agent providers; adoption of 
\texttt{agents.txt} by publishers as a 
machine-readable access policy standard; 
and expansion of \texttt{llms.txt} 
\cite{llmstxt2024} from a content permission 
declaration into a richer capability 
declaration incorporating access terms and 
supervision levels. These measures require 
no new standards bodies and no new 
infrastructure — they are extensions of 
existing HTTP and web conventions that 
individual actors can adopt unilaterally.

\textbf{Phase 2 — Economic and Format 
Infrastructure (medium-term).} Once 
identification and declaration are in place, 
the economic and content format 
infrastructure can be built. This includes: 
ATML as a parallel content delivery format 
alongside HTML; OAuth delegation tokens for 
subscription inheritance; token-based billing 
APIs for content publishers; and pilot 
implementations of the commissioned content 
economy model. These measures require 
coordination between AI providers, content 
publishers, and payment infrastructure 
operators — achievable through industry 
working groups without formal standards 
bodies.

\textbf{Phase 3 — Full Agent-First 
Infrastructure (longer-term).} The final 
phase establishes ATML as the primary 
agent-facing content format; deploys full 
cryptographic provenance chain 
infrastructure built on C2PA 
\cite{c2pa2024}; launches agent-optimized 
search indices; and establishes the agent 
advertising protocol. These measures 
require formal standardization through 
bodies such as the W3C and IETF, analogous 
to the standardization processes that 
established HTTPS \cite{felt2017measuring} 
and OAuth \cite{hardt2012oauth}.

Table~\ref{tab:migration} summarizes the 
migration roadmap.

\begin{table}[H]
\centering
\caption{Agent-first web migration roadmap.}
\label{tab:migration}
\begin{adjustbox}{width=0.48\textwidth}
\begin{tabular}{p{1.2cm}p{2.2cm}p{2.2cm}p{1.8cm}}
\toprule
\thead{Phase} &
\thead{Focus} &
\thead{Key proposals} &
\thead{Coordination\\required} \\
\midrule
\textbf{1} &
Identification and declaration &
Agent metadata headers, agents.txt, llms.txt expansion &
Unilateral — individual actors \\
\addlinespace
\textbf{2} &
Economic and format infrastructure &
ATML, OAuth delegation, token billing &
Industry working groups \\
\addlinespace
\textbf{3} &
Full agent-first infrastructure &
Provenance chain, agent search, ad protocol &
Formal standards bodies \\
\bottomrule
\end{tabular}
\end{adjustbox}
\end{table}

\subsection{Backward Compatibility}
\label{sec:framework:backward}

A recurring concern with proposals to 
redesign web infrastructure is disruption 
to the existing ecosystem. We emphasize 
that every proposal in this framework is 
\textit{additive rather than disruptive}. 
No existing web standard is deprecated. 
No existing content format is invalidated. 
No existing economic model is prohibited.

Agent metadata headers are additional HTTP 
headers — servers that do not recognize 
them ignore them. \texttt{agents.txt} is 
an additional well-known file — sites that 
do not provide it simply have no declared 
agent policy, and agents fall back to 
default behaviors. ATML is an additional 
content type — browsers that do not 
request it receive HTML as before. Token 
billing is an additional payment option — 
publishers who do not adopt it continue 
to operate under existing models.

The human web does not disappear in the 
agent-first future — it becomes one layer 
of a richer, dual-layer web that serves 
both human and agent clients from the same 
infrastructure. This is not a replacement; 
it is an extension. The web has extended 
itself before — adding HTTPS alongside 
HTTP, adding mobile layouts alongside 
desktop, adding APIs alongside HTML pages 
— and has been strengthened by each 
extension. The agent-first layer is the 
next extension in this lineage.

\subsection{Relationship to Existing 
             Protocol Work}
\label{sec:framework:protocols}

The agent-first internet framework proposed 
in this paper is complementary to, not 
competitive with, the protocol-level work 
surveyed in Section~\ref{sec:related}. MCP 
\cite{anthropic2024mcp}, A2A 
\cite{google2025a2a}, NLWeb 
\cite{microsoft2025nlweb}, and related 
protocols address agent-to-tool and 
agent-to-agent communication — the 
plumbing of the agentic web. The 
agent-first internet framework addresses 
the social contract, economic model, and 
content architecture of the web that 
these agents operate within — the 
foundations those protocols build on.

The relationship is analogous to the 
relationship between TCP/IP and the web's 
application layer: TCP/IP provides the 
communication substrate; HTTP, HTML, and 
the web's economic conventions constitute 
the application layer built on top. MCP 
and A2A are the emerging TCP/IP of the 
agentic web; the agent-first internet 
framework is the application layer — the 
conventions, economic models, and content 
standards that determine what agents 
actually do with the communication 
capabilities those protocols provide.

\subsection{Summary}
\label{sec:framework:summary}

The agent-first internet framework 
synthesizes fifteen specific proposals 
across three layers into a coherent 
redesign grounded in a single philosophical 
anchor — the agent-as-human-proxy principle 
— and expressed as ten design principles 
that practitioners can apply when making 
architectural decisions about agent-web 
interaction. The framework is additive, 
backward-compatible, and phased — 
preserving the human web throughout the 
transition to an agent-first architecture. 
It positions the challenge not as a 
technical problem of protocol design but 
as a sociotechnical problem requiring 
renegotiation of the web's foundational 
social contract — a renegotiation that 
the web has successfully accomplished 
before and must accomplish again.

\section{Open Challenges}
\label{sec:challenges}

\begin{figure}[H]
\centering
\includegraphics[width=\textwidth]{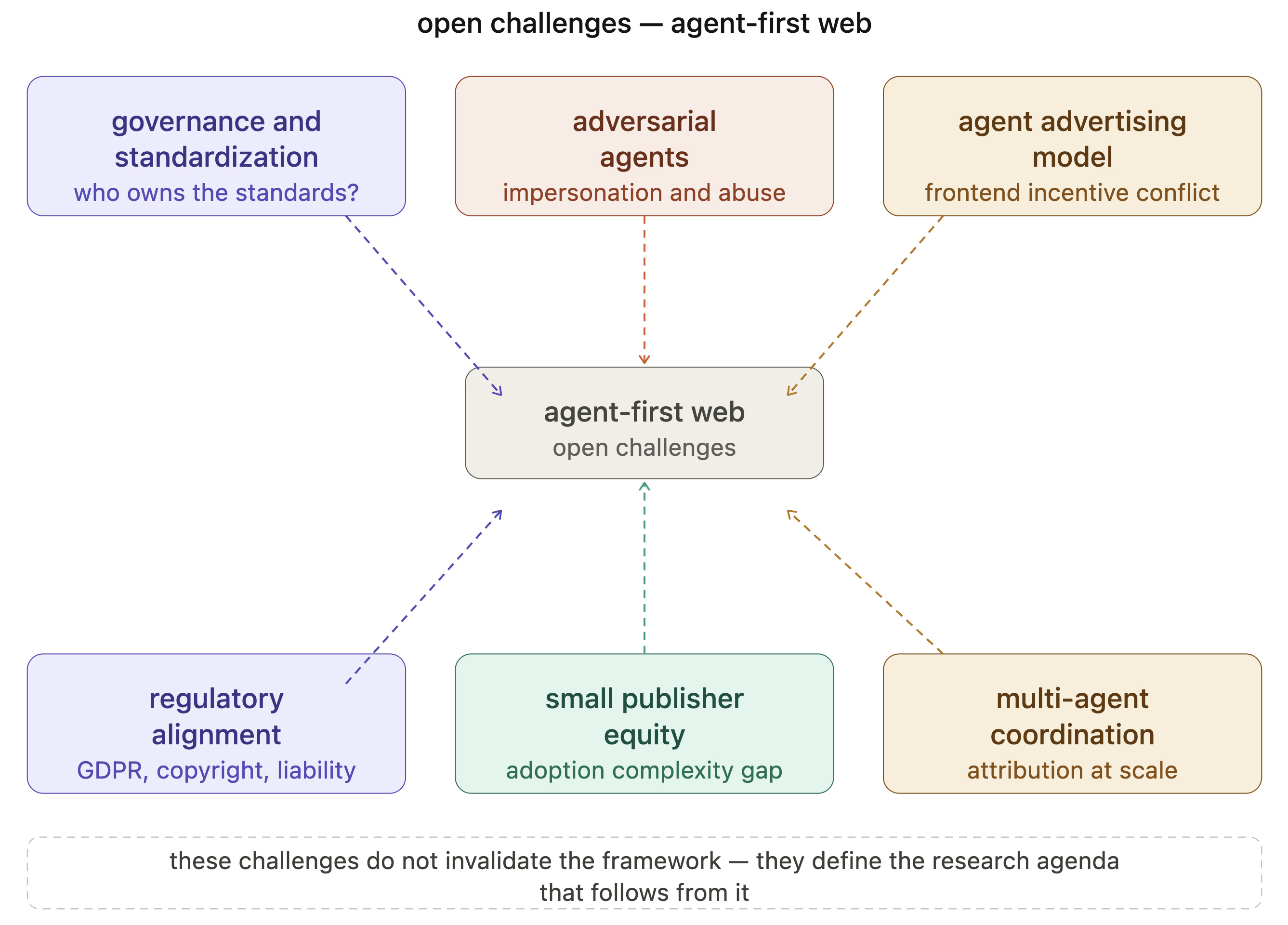}
\caption{Open challenges in the agent-first web. 
Six challenge areas remain unresolved by the 
proposed framework: governance and standardization, 
adversarial agents, the agent advertising model, 
regulatory alignment, small publisher equity, and 
multi-agent coordination. These challenges do not 
invalidate the framework — they define the research 
agenda that follows from it.}
\label{fig:challenges}
\end{figure}

The agent-first internet framework proposed in 
this paper is principled and internally coherent, 
but it does not resolve all problems. Honest 
accounting of what the framework does not solve 
is as important as what it proposes — it defines 
the research agenda that follows from this work 
and prevents overreach in the claims made. This 
section identifies six open challenges that 
remain unresolved, each constituting a direction 
for future work. Figure~\ref{fig:challenges} 
provides an overview.

\subsection{Governance and Standardization}
\label{sec:challenges:governance}

The proposals in this paper — agent metadata 
headers, \texttt{agents.txt}, ATML, the 
supervision tier model, the provenance chain 
architecture — require standardization to achieve 
the universal adoption that makes them effective. 
A metadata header that only some agent providers 
implement provides partial identification at best. 
A content format that only some publishers adopt 
creates a two-tier web rather than a unified 
agent-first layer. Standards achieve their value 
through universality, and universality requires 
a governance process~\cite{think-before-act}.

The web's history of standardization is instructive 
but not entirely encouraging. HTML was standardized 
through the W3C, a process that produced robust 
but slow-moving standards vulnerable to 
browser-vendor fragmentation during the browser 
wars of the late 1990s \cite{raggett1999html}. 
HTTP was standardized through the IETF, a more 
technically rigorous but equally slow process. 
OAuth emerged from industry collaboration before 
IETF formalization \cite{hardt2012oauth}. Each 
pathway involved years of negotiation between 
parties with conflicting commercial interests.

The agent-first web faces a governance challenge 
of comparable complexity, with the additional 
complication that the primary stakeholders — AI 
companies, content publishers, infrastructure 
providers, and regulators — have interests that 
are not merely competitive but structurally 
opposed. AI companies benefit from open agent 
access; publishers benefit from controlled, 
economically compensated access; infrastructure 
providers benefit from being the chokepoint of 
any access control mechanism; regulators benefit 
from accountability that none of the other 
parties are inclined to provide voluntarily.

We do not propose a governance solution here — 
the political economy of web standards governance 
is a research domain in its own right. We note 
that the Phase 1 proposals in our migration 
roadmap (Section~\ref{sec:framework:migration}) 
are deliberately designed to be adoptable 
unilaterally by individual actors without 
standards body coordination, precisely to 
create momentum and demonstrated value before 
formal standardization is required.

\subsection{The Adversarial Agent Problem}
\label{sec:challenges:adversarial}

The agent identification metadata mechanism 
proposed in Section~\ref{sec:access:metadata} 
assumes that agent metadata declarations are 
honest. The Perplexity case demonstrated that 
this assumption does not hold under adversarial 
conditions: bad actors will modify agent 
identifiers, rotate IP addresses, and falsify 
intent declarations to circumvent access controls 
\cite{perplexity2024sharing}. Cryptographic 
signing of agent metadata raises the cost of 
impersonation significantly but does not 
eliminate it — a bad actor who controls their 
own signing key can still declare false intent, 
and a compromised legitimate signing key can 
be used for impersonation until revocation.

The adversarial agent problem is structurally 
similar to the email spam problem. Email 
authentication standards — SPF, DKIM, DMARC — 
significantly reduced certain classes of email 
abuse by making sender identity verifiable, but 
did not eliminate spam or phishing; they shifted 
the attack surface rather than eliminating it 
\cite{allman2007dkim}. Agent metadata 
authentication is likely to follow a similar 
trajectory: reducing casual circumvention while 
displacing sophisticated attackers toward more 
costly evasion strategies.

A more robust long-term defense is behavioral 
rather than declarative: monitoring agent 
consumption patterns over time to detect 
statistical anomalies — consumption rates 
inconsistent with declared personal use, 
content extraction patterns consistent with 
training rather than query-answering, access 
patterns consistent with competitive 
intelligence rather than user assistance. 
Behavioral detection is harder to evade than 
declaration forgery, but requires infrastructure 
that does not currently exist at web scale and 
raises privacy concerns about agent activity 
monitoring that must be carefully managed.

\subsection{The Agent Advertising Model}
\label{sec:challenges:advertising}

The agent advertising model sketched in 
Section~\ref{sec:economics:advertising} faces 
a fundamental principal-agent conflict that 
the framework does not resolve. Publishers 
want their advertisements surfaced to human 
users by agent frontends. Agent frontends — 
Claude, ChatGPT, Gemini — have competing 
monetization interests and limited incentive 
to surface publisher advertisements that 
generate no revenue for the frontend operator. 
Users want clean, uninterrupted agent 
responses and are likely to prefer frontends 
that do not surface advertisements.

This three-way conflict cannot be resolved 
by technical standards alone. Two resolution 
pathways exist, both of which involve 
stakeholder coordination beyond what any 
single actor can achieve. The first is a 
revenue-sharing model in which publisher 
advertisement revenue is split between the 
publisher and the agent frontend operator 
that surfaces it — giving frontend operators 
a financial incentive to carry publisher 
advertisements. The second is a regulatory 
mandate analogous to must-carry rules in 
broadcast television regulation, requiring 
agent frontends to surface publisher 
advertisements as a condition of accessing 
publisher content.

Neither pathway is politically straightforward. 
Revenue sharing requires bilateral agreements 
between every publisher and every agent 
frontend operator — a combinatorially complex 
negotiation landscape. Regulatory mandates 
require jurisdictional agreement across 
regulators who currently lack frameworks 
for agent content intermediation. We identify 
this as a priority open challenge and flag 
it as a direction for future work in both 
the technical and policy research communities.

\subsection{Regulatory Alignment}
\label{sec:challenges:regulatory}

The existing regulatory landscape for web 
content and data was designed for human-web 
interaction and applies imperfectly or 
ambiguously to agent-mediated interaction 
across multiple dimensions.

\textbf{Copyright.} When an AI agent retrieves 
and synthesizes content from multiple sources, 
the copyright status of the synthesized output 
is legally unsettled in most jurisdictions. 
The training data copyright disputes currently 
before courts in the United States and European 
Union \cite{henderson2023foundation} address 
a related but distinct question — whether 
training on copyrighted data constitutes 
infringement — and their resolution will not 
definitively answer the question of whether 
agent synthesis of copyrighted content for 
a user constitutes fair use or requires 
licensing. The token-based licensing model 
proposed in Section~\ref{sec:economics:tokens} 
is designed to be compatible with whatever 
copyright resolution emerges, but cannot 
itself produce that resolution.

\textbf{Data protection.} GDPR and equivalent 
regulations govern the processing of personal 
data by automated systems, with requirements 
around consent, purpose limitation, and data 
subject rights \cite{gdpr2016}. When an AI 
agent processes content containing personal 
data on behalf of a user, it is unclear which 
entity — the user, the agent provider, the 
content publisher — bears data controller 
responsibility. The delegation token mechanism 
proposed in Section~\ref{sec:access:subscription} 
partially addresses this by establishing a 
clear authorization chain, but does not resolve 
the underlying regulatory ambiguity.

\textbf{Platform liability.} Section 230 of 
the US Communications Decency Act and 
equivalent provisions in other jurisdictions 
limit platform liability for user-generated 
content. It is unclear whether AI agents that 
synthesize and present content to users are 
acting as platforms (potentially eligible for 
liability protection) or as publishers 
(potentially liable for the content they 
synthesize). This ambiguity has significant 
implications for the design of agent content 
systems and requires legislative rather than 
technical resolution.

\subsection{The Small Publisher Equity Problem}
\label{sec:challenges:equity}

The proposals in this paper — ATML publishing 
infrastructure, provenance chain certification, 
token billing APIs, \texttt{agents.json} 
capability declarations — carry implementation 
costs that large publishers can absorb and 
small publishers cannot. The New York Times 
has engineering teams capable of implementing 
dual-layer publishing infrastructure. An 
independent journalist with a WordPress blog 
does not.

If the agent-first web is accessible only to 
publishers with the resources to implement its 
infrastructure, it creates a two-tier web in 
which large publishers dominate agent-accessible 
content and small, independent, and community 
publishers are effectively excluded. This 
replicates and potentially amplifies existing 
web inequality — large platforms already 
dominate search rankings and social distribution; 
agent-first infrastructure could extend this 
dominance to agent-mediated information access.

The mitigation is platform-level adoption: 
if WordPress, Substack, Ghost, and similar 
publishing platforms implement ATML generation, 
\texttt{agents.json} capability declarations, 
and provenance tagging as platform features, 
their users gain agent-first capabilities 
without individual implementation effort. 
This mirrors how SSL adoption was democratized 
through hosting platform support rather than 
individual webmaster implementation. We 
recommend that the standards proposed in 
this paper be designed with platform-level 
implementation as the primary deployment 
pathway — simplicity for platform integration 
is more important than feature richness 
for individual deployment.

\subsection{Multi-Agent Coordination}
\label{sec:challenges:multiagent}

The framework proposed in this paper addresses 
the interaction between a single agent and web 
content. Increasingly, however, agent tasks 
involve multiple agents collaborating — an 
orchestrator agent directing specialist agents 
to retrieve, analyze, and synthesize content 
from many sources simultaneously 
\cite{durante2024agent}. This multi-agent 
scenario introduces attribution and billing 
complexities that the framework does not resolve.

When five specialist agents each retrieve 
content from ten sources to complete a single 
user task, who pays for the content access? 
The user, through their orchestrating agent? 
Each specialist agent independently? The 
orchestrator on behalf of all specialists? 
The token-based billing model proposed in 
Section~\ref{sec:economics:tokens} provides 
a unit of account but not a billing 
architecture for multi-agent scenarios.

Attribution is equally complex. When a 
multi-agent pipeline produces a synthesized 
output that incorporates content from fifty 
sources retrieved by multiple agents, the 
provenance chain described in 
Section~\ref{sec:content:provenance} must 
span agent boundaries — tracking not just 
which sources were consulted but which agents 
consulted them and how their outputs were 
combined. This requires a cross-agent 
provenance standard that does not currently 
exist and whose design involves significant 
technical and privacy tradeoffs.

The A2A protocol \cite{google2025a2a} and 
related multi-agent communication standards 
provide communication infrastructure for 
multi-agent systems but do not address 
content attribution or billing coordination. 
We identify multi-agent content economics 
as a priority area for future work, 
noting that its resolution requires 
coordination between the agent communication 
protocol community and the web content 
economics community — two communities that 
have not yet engaged substantively with 
each other's work.

\subsection{Summary}
\label{sec:challenges:summary}

Table~\ref{tab:challenges} summarizes the 
six open challenges and their relationship 
to the framework proposals.

\begin{table}[H]
\centering
\caption{Open challenges and their relationship 
to the agent-first web framework.}
\label{tab:challenges}
\begin{adjustbox}{width=0.48\textwidth}
\begin{tabular}{p{2.2cm}p{2.2cm}p{2.4cm}}
\toprule
\thead{Challenge} &
\thead{Framework\\limitation} &
\thead{Resolution\\pathway} \\
\midrule
Governance and standardization &
Proposals require universal adoption &
W3C, IETF, industry working groups \\
\addlinespace
Adversarial agents &
Declarations can be falsified &
Behavioral detection + cryptographic signing \\
\addlinespace
Agent advertising &
Frontend incentive conflict &
Revenue sharing or regulatory mandate \\
\addlinespace
Regulatory alignment &
Legal frameworks predate agents &
Legislative and judicial resolution \\
\addlinespace
Small publisher equity &
Implementation costs create barriers &
Platform-level adoption pathway \\
\addlinespace
Multi-agent coordination &
Single-agent billing and attribution model &
Cross-agent provenance standard \\
\bottomrule
\end{tabular}
\end{adjustbox}
\end{table}

These six challenges are significant but 
do not invalidate the framework. Each 
represents a known limitation with an 
identifiable resolution pathway — they 
are open problems, not fatal flaws. The 
web has faced comparable challenges at 
every major architectural transition: 
HTTPS deployment faced governance 
fragmentation and small-site adoption 
barriers before Let's Encrypt democratized 
certificate issuance; OAuth faced 
competing implementations before RFC 
6749 standardized the protocol; mobile 
web faced regulatory uncertainty before 
jurisdictions developed mobile-specific 
frameworks. The agent-first web transition 
will follow a similar pattern — early 
adoption by large actors, gradual 
standardization, eventual universal 
deployment — provided the foundational 
framework is principled enough to 
survive the transition intact. We 
believe the framework proposed in this 
paper meets that standard.

\section{Conclusion}
\label{sec:conclusion}

The World Wide Web was built for human eyes. For 
three decades, this assumption was so fundamental 
that it required no articulation — it was simply 
the water in which the web swam. The emergence of 
AI agents as primary intermediaries between humans 
and web content has made this assumption visible 
by breaking it. Blanket agent blocking, collapsing 
publisher revenues, and the epistemic recursion 
loop are not isolated problems — they are 
symptoms of a single underlying condition: a web 
whose architecture was designed for a world that 
no longer exists.

This paper has argued that the response to this 
condition cannot be reactive or piecemeal. 
Patching the access layer without the economic 
layer produces agents that can reach content but 
publishers who cannot sustain producing it. 
Patching the economic layer without the content 
layer produces billing infrastructure for content 
that degrades in quality through epistemic 
recursion. The three layers — access, economics, 
and content — are deeply coupled, and must be 
redesigned simultaneously, grounded in a single 
philosophical anchor: that AI agents acting on 
behalf of humans are first-class web citizens 
entitled to the same presumption of access, 
subject to equivalent obligations, and deserving 
of an architectural environment designed for 
their interaction model.

The agent-first internet framework proposed in 
this paper offers that redesign. At the access 
layer, agent identification metadata headers and 
a graduated rate-limiting model replace blanket 
blocking with context-appropriate access control. 
At the economic layer, the intent-based tier 
model and token-based subscription mechanism 
replace the broken attention economy with a 
value-exchange model grounded in the 
agent-as-human-proxy principle. At the content 
layer, ATML, the human supervision tier model, 
and the provenance chain architecture replace 
opaque HTML with semantically rich, 
provenance-anchored content that breaks the 
epistemic recursion loop. Together these 
proposals constitute ten design principles for 
the agent-first internet — a framework intended 
to serve practitioners making architectural 
decisions about agent-web interaction in the 
same way that REST principles served API 
designers and mobile-first principles served 
responsive web designers.

The web has renegotiated its social contract 
before. The transition from HTTP to HTTPS 
reframed security from an optional feature to 
a default expectation. The transition from 
desktop to mobile reframed layout from a fixed 
assumption to a responsive variable. Each 
transition required years of friction, partial 
adoption, and eventual standardization before 
becoming universal. The transition to an 
agent-first web is the next renegotiation in 
this lineage — and by the evidence of collapsing 
publisher revenues, infrastructure-scale agent 
blocking, and accelerating epistemic recursion, 
it is already overdue.

The cost of inaction is not merely economic. 
A web that successfully excludes AI agents 
excludes the humans those agents represent — 
making the web less useful to the people it 
was built to serve. A web whose content is 
generated by AI for consumption by AI, without 
human intentionality at any point in the chain, 
is no longer a record of human knowledge — it 
is a hall of mirrors, reflecting increasingly 
distorted versions of what humans once knew. 
Redesigning the web for AI agents is not a 
concession to technological change. It is an 
act of stewardship for the knowledge ecosystem 
that both humans and AI depend upon.



\bibliographystyle{elsarticle-harv}
\bibliography{reference}

\end{document}